%% file: amorsyn.tex
\documentclass{article}

\PassOptionsToPackage{numbers}{natbib}

\usepackage[final]{neurips_2021}

\usepackage[utf8]{inputenc} 
\usepackage[T1]{fontenc}    
\usepackage{hyperref}       
\usepackage{url}            
\usepackage{booktabs}       
\usepackage{amsfonts}       
\usepackage{nicefrac}       
\usepackage{microtype}      
\usepackage{xcolor}         

\input{packages}
\input{macros}

\title{Amortized Synthesis of Constrained Configurations Using a Differentiable Surrogate}

\author{
    Xingyuan Sun$^1$,\, Tianju Xue$^2$,\, Szymon Rusinkiewicz$^1$,\, Ryan P. Adams$^1$ \\
    $^1$Department of Computer Science\quad$^2$Department of Civil and Environmental Engineering \\
    Princeton University \\
    \texttt{\{xs5, txue, smr, rpa\}@princeton.edu} \\
}

\begin{document}

\maketitle

\input{text/abstract}
\input{text/intro}
\input{text/related_work}
\input{text/method}
\input{text/pt}
\input{text/rb}
\input{text/discussion}

\input{text/acknowledgements}

\bibliography{amorsyn}
\bibliographystyle{plainnat}


\newpage
\appendix
\input{text/appendix-pt}
\input{text/appendix-rb}

\end{document}

%% file: macros.tex
\newcolumntype{L}[1]{>{\raggedright\let\newline\\\arraybackslash\hspace{0pt}}m{#1}}
\newcolumntype{C}[1]{>{\centering\let\newline\\\arraybackslash\hspace{0pt}}m{#1}}
\newcolumntype{R}[1]{>{\raggedleft\let\newline\\\arraybackslash\hspace{0pt}}m{#1}}

\newcommand{\sect}[1]{Section~\ref{#1}}
\newcommand{\ssect}[1]{\S~\ref{#1}}
\newcommand{\eqn}[1]{Equation~\ref{#1}}

\newcommand{\fig}[1]{Figure~\ref{#1}}

\newcommand{\tbl}[1]{Table~\ref{#1}}

\newcommand{\degree}{\ensuremath{^\circ}\xspace}
\newcommand{\ignore}[1]{}
\newcommand{\norm}[1]{\lVert#1\rVert}

\makeatletter
\DeclareRobustCommand\onedot{\futurelet\@let@token\@onedot}
\def\@onedot{\ifx\@let@token.\else.\null\fi\xspace}

\def\eg{\emph{e.g}\onedot} 
\def\ie{\emph{i.e}\onedot} 
 
\def\etc{\emph{etc}\onedot}

\makeatother

\definecolor{MyDarkBlue}{rgb}{0,0.08,1}
\definecolor{MyDarkGreen}{rgb}{0.02,0.6,0.02}
\definecolor{MyDarkRed}{rgb}{0.8,0.02,0.02}
\definecolor{MyDarkOrange}{rgb}{0.40,0.2,0.02}
\definecolor{MyPurple}{RGB}{111,0,255}
\definecolor{MyRed}{rgb}{1.0,0.0,0.0}
\definecolor{MyGold}{rgb}{0.75,0.6,0.12}
\definecolor{MyDarkgray}{rgb}{0.66, 0.66, 0.66}

\makeatletter
\def\@squeeze{0.44}

\newlength\@myskip
\ifdim \@myskip < \parskip \@myskip=\parskip \fi
\advance\@myskip-\@squeeze\@myskip
\advance\parskip-\@squeeze\parskip
\newlength\@myextrastretch \@myextrastretch=\@myskip
\newdimen\@mytempdimen \@mytempdimen=\@myskip
\advance\@myextrastretch-\@mytempdimen

\let\@originalnormalsize=\normalsize
\renewcommand\normalsize{%
        \@originalnormalsize
        \abovedisplayskip\@myskip
        \abovedisplayshortskip 0pt plus \@myextrastretch
        \belowdisplayskip\abovedisplayskip
        \belowdisplayshortskip\belowdisplayskip
}
\normalsize

\newlength\secskipi   \secskipi=1.728\@myskip \advance\secskipi-\parskip
\newlength\secskipii  \secskipii=1.44\@myskip \advance\secskipii-\parskip 
\newlength\secskipiii \secskipiii=1.2\@myskip \advance\secskipiii-\parskip
\newlength\secskipiv  \secskipiv=1.01\@myskip \advance\secskipiv-\parskip
\advance\secskipi\@myextrastretch \advance\secskipii\@myextrastretch
\advance\secskipiii\@myextrastretch \advance\secskipiv\@myextrastretch
\renewcommand\section{\@startsection{section}{1}{\z@}%
                {-\secskipi}%
                {\secskipii}%
                {\reset@font\large\bfseries}}
\renewcommand\subsection{\@startsection{subsection}{2}{\z@}%
                {-\secskipii}%
                {\secskipiii}%
                {\reset@font\normalsize\bfseries}}
\renewcommand\subsubsection{\@startsection{subsubsection}{3}{\z@}%
                {-\secskipiii}%
                {\secskipiv}%
                {\reset@font\normalsize\bfseries}}
\renewcommand\paragraph{\@startsection{paragraph}{4}{\z@}%
                {\secskipiv}%
                {-1.0em}%
                {\reset@font\normalsize\bfseries}}
                
\begingroup\lccode`\~=`\_\lowercase{\endgroup\def~{\futurelet\next\s@@b}}
\def\s@@b{\ifcat\relax\noexpand\next\expandafter\sb\else
 \expandafter\s@@b@\fi}
\def\s@@b@#1{\sb{\futurelet\next\sb@#1}}
\def\sb@{%
 \ifx\next\space@\def\next@. {\futurelet\next\sb@}\else
  \def\next@.{%
   \ifx\next f\mkern-\thr@@ mu\else
   \ifx\next j\mkern-\thr@@ mu\else
   \ifx\next p\mkern-\tw@ mu\else
   \ifx\next t\mkern\@ne mu\else
   \ifx\next y\mkern-\@ne mu\else
   \ifx\next A\mkern-\tw@ mu\else
   \ifx\next B\mkern-\@ne mu\else
   \ifx\next D\mkern-\@ne mu\else
   \ifx\next H\mkern-\@ne mu\else
   \ifx\next I\mkern-\@ne mu\else
   \ifx\next K\mkern-\@ne mu\else
   \ifx\next L\mkern-\@ne mu\else
   \ifx\next M\mkern-\@ne mu\else
   \ifx\next P\mkern-\@ne mu\else
   \ifx\next X\mkern-\tw@ mu\else
   \fi\fi\fi\fi\fi\fi\fi\fi\fi\fi\fi\fi\fi\fi\fi}%
 \fi
 \next@.}
\mathcode`\_=\string"8000

\normalfont\normalsize
\calculate@math@sizes

\makeatother

%% file: text/abstract.tex
\begin{abstract}

In design, fabrication, and control problems, we are often faced with the task of \emph{synthesis}, in which we must generate an object or configuration that satisfies a set of constraints while maximizing one or more objective functions.
The synthesis problem is typically characterized by a physical process in which many different realizations may achieve the goal.
This many-to-one map presents challenges to the supervised learning of feed-forward synthesis, as the set of viable designs may have a complex structure.
In addition, the non-differentiable nature of many physical simulations prevents efficient direct optimization.
We address both of these problems with a two-stage neural network architecture that we may consider to be an autoencoder.
We first learn the decoder: a differentiable surrogate that approximates the many-to-one physical realization process.
We then learn the encoder, which maps from goal to design, while using the fixed decoder to evaluate the quality of the realization.
We evaluate the approach on two case studies: extruder path planning in additive manufacturing and constrained soft robot inverse kinematics.
We compare our approach to direct optimization of the design using the learned surrogate, and to supervised learning of the synthesis problem.
We find that our approach produces higher quality solutions than supervised learning, while being competitive in quality with direct optimization, at a greatly reduced computational cost.

\end{abstract}

%% file: text/intro.tex
\section{Introduction}
\label{sec:intro}

One of the ambitions of artificial intelligence is to automate problems in design, fabrication, and control that
demand efficient and accurate interfaces between machine learning algorithms and physical systems.
Whether it is optimizing the topology of a mechanical structure or identifying the feasible paths for a manufacturing robot, we can often view these problems through the lens of \emph{synthesis}.
In a synthesis task, we seek configurations of a physical system that achieve certain desiderata while satisfying given constraints; \ie, we must optimize a physically-realizable design.

In this work, \emph{design} refers to the parametric space over which we have control and in which, \emph{e.g.}, we optimize.
A \emph{realization} is the object that arises when the design is instantiated, while
\emph{goal} refers to its desired properties.
For example, in fabrication, the design might be a set of assembly steps, the realization would be the resulting object, while the goal could be to match target dimensions while maximizing strength.
Synthesis, then, refers to finding a design whose realization achieves the goal.

Synthesis problems are challenging for several reasons.
The physical realization process may be costly and time-consuming, making evaluation of many designs difficult.
Moreover, the realization process---or even a simulation of it---is generally not differentiable, rendering efficient gradient-based methods inapplicable.
Finally, there may be a many-to-one map from the parametric space of feasible and equally-desirable designs to realizations;
\ie, there may be multiple ways to achieve the goal.

\input{figText/teaser}

Surrogate modeling is widely used to address the first two challenges, though it can still be expensive because of the need for optimization, sampling, or search algorithms to find a feasible design.
More seriously, the third challenge---lack of uniqueness---creates difficulties for na\"ive supervised learning approaches to synthesis.
Specifically, consider generating many design/realization pairs, evaluating the constraints and objectives on the realizations, and attempting to learn a supervised map from goal back to design.
When multiple designs lead to the same realization, or multiple realizations achieve the same goal, the supervised learner is penalized for producing designs that are valid but happen to not be the ones used to generate the data.
Moreover, this approach may learn to produce an ``average'' design that is actually incorrect.
Figure~\ref{fig:teaser}a shows a cartoon example: if the goal is to throw a ball to reach some target distance,
there are two possible launch angles (designs) resulting in landing points (realizations) at the correct spot.
Performing least-squares regression from distance to angle on the full set of distance/angle pairs, however, learns an average angle that does not satisfy the goal.

To address these challenges, we propose to use a two-stage neural network architecture that resembles an autoencoder.
One stage (the decoder) acts as a differentiable surrogate capturing the many-to-one physical realization process.
The other stage (the encoder) maps from a goal back to a design but, critically, it is trained end-to-end with a loss in the space of realizations that flows back through the decoder.
Thus the encoder---our central object of interest for synthesis---is not constrained to match a \emph{specific} design in a training dataset, but instead is tasked with finding \emph{any} design that meets the desiderata of the realized output.
The result is a neural network that performs \emph{amortized} synthesis: it is trained once, and at run time produces a design that is approximately optimal, using only a feed-forward architecture.
Note that our method is not an autoencoder, as the design is not a lower-dimensional representation of the goal, and the encoder and the decoder are trained in separate stages.

Our method places a number of requirements on the synthesis problem.
First, to train the surrogate, we need data pairs of designs and realizations. Commonly, this would require us to generate designs, in which a substantial amount of designs are viable, and simulate them on a simulator.
Second, given our current setting, the physical realization process needs to be deterministic.
Third, to train the encoder, the synthesis problem should have a clear objective function, or at least we can quantify the objective.
Finally, given our current feed-forward setting, we consider synthesis problems that need only one valid design, although we may further extend our method by using a generative encoder.

In this work, we demonstrate this two-stage approach on a pair of specific design tasks.
The first case study is extruder path planning for a class of 3D printers (the Markforged Mark Two) that can reinforce polymer layers with discrete fibers (\fig{fig:teaser}b).
Since the fibers are stiff, their shape is deformed after extrusion (\fig{fig:teaser}c, top row),
and our task is to find an extruder path that results in a given fiber shape.
As shown in~\fig{fig:teaser}c, this problem has the many-to-one nature described above: for a small error tolerance on fiber path, there exist infinitely many extruder paths, which may even look very different.
The second case study is constrained soft robot inverse kinematics.
In this work, we use a simulation of a snake-like soft robot as in~\citet{xue2020amortized}, in which we can control the stretch ratios of each individual segment on both sides of the robot.
The robot has to reach a target while avoiding an obstacle, and the locations of both are input goals.
As before, there may be multiple solutions for a given goal (\ie, locations of target and obstacle), as shown in~\fig{fig:teaser}d.

For both case studies, we compare to two baseline algorithms. In \texttt{direct-learning}, a neural network for the synthesis problem (\ie, from goal to design) is trained in a supervised manner on a set of designs. Since this effectively averages designs in the training dataset, as argued above, our method outperforms it significantly. The second baseline is \texttt{direct-optimization}, which uses a gradient-based method (BFGS) to optimize for each new design separately, given access to the trained differentiable surrogate for the realization process (decoder). Our method is competitive with this rough ``performance upper bound'' while using dramatically lower computational resources.

%% file: figText/teaser.tex
\begin{figure}[t]
    \centering
    \includegraphics[width=\linewidth]{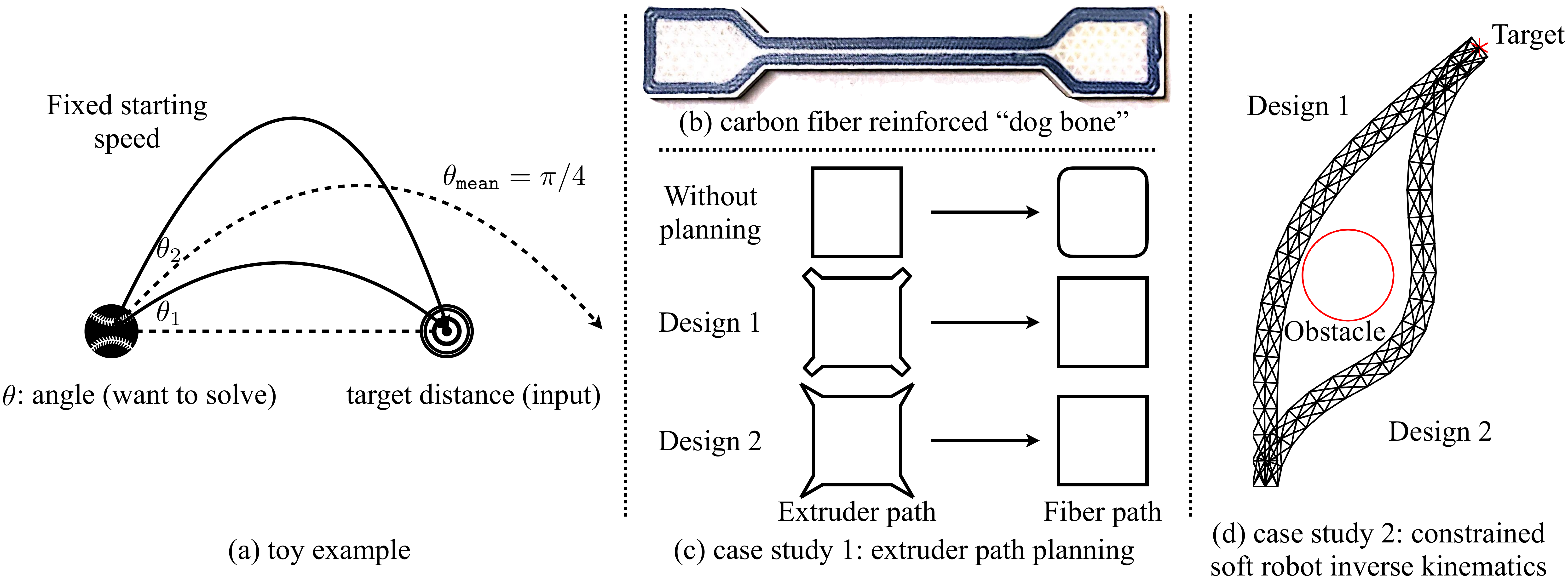}
    \caption{(a) For a fixed and large-enough starting speed, there exist exactly two angles such that the ball will hit the target, where the mean of these two angles is $\pi / 4$. (b) Some 3D printers utilize fibers to reinforce the thermoplastic print.  (c) For such printers, fiber is laid out along an extruder path but deforms into a smoothed version due to the fiber's high stiffness and low stretch. Our goal is to generate extruder paths that compensate for the smoothing, but multiple extruder paths can result in the same target shape, such as a square. (d) In soft robot inverse kinematics, we control the stretch ratios of both the left- and right-hand sides of a snake-like robot.  Our goal is to reach the target while avoiding an obstacle but, as is illustrated, the solution is not unique -- two different designs are shown.}
    \label{fig:teaser}
\end{figure}

%% file: text/related_work.tex
\section{Related work}
\label{sec:related_work}

\paragraph{Machine learning applications in synthesis problems.}
In synthesis tasks, the aim is to find a design solution such that its realization achieves one or more given goals. Usually, the solution is non-unique, and only one is needed.
In molecule discovery, one would like to find a molecule which has some desired properties (\eg, minimal side effects, efficacy, metabolic stability, growth inhibition)~\citep{coley2018machine, jin2020multi, polykovskiy2018entangled, hawkins2021generating, stokes2020deep}. See~\citet{vamathevan2019applications, chen2018rise} for surveys on machine/deep learning in drug discovery. Challenges of molecule discovery include discrete design space~\citep{gomez2018automatic, NEURIPS2019_febefe1c} and limited data~\citep{jin2021deep} due to difficulty in simulation.
In materials synthesis, researchers seek materials with specific properties. See~\citet{bhuvaneswari2021deep} for a review. Similarly, to obtain enough training data, researchers have put efforts into parsing and learning from scientific literature in natural language~\citep{kim2017materials, huo2019semi}.
In 3D shape generation, one wants to find a 3D shape that has some desired properties: properties like 2.5D sketches~\citep{marrnet} that can be easily calculated and properties like functionality~\citep{guan2020fame} that need to be human-labeled.
In topology optimization, researchers aim to maximize the system's performance by optimizing the material layout given boundary conditions, constraints, external loads, etc~\citep{sigmund2013topology, bendsoe2013topology, wang2003level}. Machine learning has been used to infer properties~\citep{sasaki2019topology}, find representations of designs~\citep{kallioras2020accelerated, yu2018deep}, and directly generate designs~\citep{kollmann2020deep}.
In program synthesis, the goal is to find programs that realize given intentions (\eg, generating 3D shapes, answering visual questions)~\citep{gulwani2014program, tian2018learning, yi2018neural}, where machine learning has been used to generate programs and execute programs.
In this work, we propose an approach to learn a feed-forward neural network that can directly and efficiently produce a feasible solution to a synthesis problem that satisfies the requirements mentioned above.

\paragraph{Surrogate/oracle-based synthesis.}
Surrogate/oracle-based synthesis uses an auxiliary model---a surrogate (or sometimes called oracle)---that can evaluate qualities of a design without time-consuming laboratory experiments, while still being reasonably accurate~\citep{koziel2011surrogate, brookes2019conditioning}. Surrogate models can be physics-based or approximation-based~\citep{koziel2014surrogate} (\ie, empirical), and there are different modeling techniques for approximation-based surrogates, including polynomial regression, radial basis functions, Gaussian processes, and neural networks~\citep{koziel2011surrogate, han2012surrogate}.
\citet{bhosekar2018advances} provide a review of surrogate-based methods, and see \citet{koziel2013surrogate} for a general review of surrogate models in engineering.
There are two major approaches: optimization and sampling.
The most common approach is surrogate-based optimization; see \citet{forrester2009recent} for a survey.
Some application examples include: optimizing the parameters of a CPU simulator \citep{renda2020difftune}; solving partial differential equations in service of PDE-constrained optimization \citep{xue2020amortized};  and optimizing stochastic non-differentiable simulators \citep{shirobokov2020black}.
Researchers have also developed methods to deal with the challenge that inputs can be out-of-distribution for the oracle~\citep{fannjiang2020autofocused, fu2020offline, trabucco2021conservative}.
On the other hand, sampling-based methods have several advantages: the design space can be discrete and can generate multiple designs~\citep{brookes2019conditioning, brookes2018design, engel2018latent}.
Researchers have successfully applied sampling methods to problems in chemistry~\citep{gomez2018automatic} and biology~\citep{gupta2019feedback, killoran2017generating, rives2021biological}.
In this work, we use surrogate-based optimization as one of our baseline algorithms, and our proposed method uses a surrogate during training to optimize for a feed-forward network for the design problem. 

\paragraph{Differentiable surrogate of losses in machine learning.}
Since some loss functions in machine learning are not differentiable (\eg, IoU for rotated bounding boxes, 0-1 loss in classification), researchers have proposed to learn surrogates for them.
\citet{grabocka2019learning} provided a formulation of surrogate loss learning and compared several learning mechanisms on some commonly used non-differentiable loss functions.
\citet{liu2020unified} proposed a general pipeline to learn surrogate losses.
\citet{bao2020calibrated, hanneke2019surrogate} provided theoretical analyses of surrogates for 0-1 loss in classification.
\citet{patel2020learning, nagendar2018neuro, yuan2020robust} explored the use of surrogate losses in various real-world tasks, including medical image classification, semantic segmentation, and text detection and recognition.
In this work, due to the non-uniqueness of design solutions, we further extend this idea from loss functions to more complex, physical realization processes.

\paragraph{Robot motion planning.}
Robot motion planning~\citep{latombe2012robot} can also be viewed as a form of synthesis, and there are different types of approaches to it.
One popular strategy is \emph{optimization}.
For example, researchers have used evolutionary algorithms (EA)~\citep{leger1999automated, cabrera2011evolutionary, hornby2001evolution}, including variants of genetic algorithms (GA)~\citep{chung1997task, chen1995determining, tabandeh2016memetic, cabrera2002optimal, khorshidi2011optimal, kim1993formulation, farritor1996systems} and covariance matrix adaptation evolution strategy (CMA-ES)~\citep{ha2016task}.
Another example is constrained optimization~\citep{whitman2018task, subramanian1995kinematic, coros2013computational, dogra2021optimal}, which includes extensions like sequential quadratic programming (SQP)~\citep{campos2020automated, ha2016task} and the DIRECT algorithm~\citep{van2009optimal}.
Other examples of optimization applied to robot motion planning include reinforcement learning~\citep{singh1994robust, everett2018motion} and teaching-learning-based optimization~\citep{sleesongsom2017four}.
Another popular class of methods is \emph{sampling}, including simulated annealing (SA)~\citep{baykal2017asymptotically, patel2015task, zhu2012motion}, probabilistic roadmaps (PRM)~\citep{karaman2011sampling}, rapidly exploring random trees (RRT)~\citep{campos2019task, baykal2017asymptotically, campos2021synthesizing}, \etc.
Finally, \emph{search} methods, \eg, A* search~\citep{ha2018computational}, have been used to solve motion planning problems.
These works usually model the robot's physics directly without using a surrogate model and solve the design using methods including optimization, sampling, and search, which can be time-consuming.
Our method amortizes the cost of inference, and can be used to solve planning problems with more complex physics that cannot be directly modeled.
In our first case study, we plan for a trajectory of the extruder, and our design space is not the speeds of the motors but rather the coordinates of points along the trajectory.
In the second case study, rather than solving a dynamic planning problem, we solve a static planning task on a soft robot, with the stretch ratios of all the controllable segments of the robot as the design space.

\paragraph{Path planning in 3D printing.}
Path planning is one of the most important problems in 3D printing, and people plan for different objectives: minimizing printing time, avoiding collision, faster planning, etc.
\citet{shembekar2018trajectory} proposed a planning algorithm to build complex shapes with multiple curvatures and can avoid collisions.
\citet{ganganath2016trajectory} minimized the printing time by modeling the task as a traveling salesman problem.
\citet{xiao2020efficient} speeded up path planning algorithms by introducing efficient topology reconstruction algorithms.
\citet{stragiotti2020continuous} provided an optimization-based algorithm that minimizes compliance of a printed part.
\citet{asif2018modelling} introduced a planning algorithm for continuous fiber and can generate a continuous deposition path.
See \citet{huang2020survey} for a review of existing work in 3D printing path design.
In this work, instead of the aforementioned objectives, we plan an extruder path to compensate for the deformation caused by the printing process. We demonstrate increased run-time efficiency by amortizing the cost of physical simulation of the printing process, learning a feed-forward network to output the extruder path.

%% file: text/method.tex
\section{Method}
\label{sec:method}

We formalize synthesis as a constrained optimization problem, denoting the set of allowed designs as~$\Theta$ and the set of possible realizations as~$\mathcal{U}$.
There is a physical process that maps from design to realization that we denote as~${U:\Theta\to\mathcal{U}}$.
Our goal may be a function of both the realization and the design, as designs may differ in, \eg, ease of manufacturing.
Moreover, it may be appropriate to specify a parametric family of goals to accommodate related tasks, \eg, different target locations in inverse kinematics.
The user expresses a ($\bm g$-indexed) family of design goals via a cost function denoted~${\mathcal{L}_{\bm g}:\Theta\times\mathcal{U}\to\mathbb{R}}$.
The problem of interest is to optimize the cost function with respect to the design:
\begin{align}
    \min_{\bm \theta\in\Theta} \mathcal{L}_{\bm g}(\bm \theta, \bm u) \quad \text{s.t.} \quad U(\bm\theta)=\bm u\,.
\end{align}
We can view this problem as a generalization of PDE-constrained optimization problems, where we have allowed for broader types of realizations than PDE solutions.
Revisiting the challenges of synthesis problems articulated earlier,~$U$ may be expensive, non-differentiable, and non-injective, and~$\mathcal{L}_{\bm g}$ may not a have a unique minimum.

Recalling the toy problem of Figure~\ref{fig:teaser}a:~$\bm \theta$ is the angle at which the ball is thrown,~${U(\bm\theta)=\bm u}$ is where it lands, the goal~$\bm g$ is a desired distance, \ie, a target realization,
and a (design-independent) cost function might be the squared difference between the desired and actual realizations:~${\mathcal{L}_{\bm g}(\bm \theta, \bm u) = ||\bm g - \bm u||^2}$.

Since the realization ${\bm u=U(\bm \theta)}$ is unique for a specific design $\bm \theta$, we propose using a two-stage approach, which can be viewed as an autoencoder. We first learn a differentiable surrogate (decoder)~$\hat{U}(\cdot)$ for the physical realization process from~$\bm \theta$ to~$\bm u$, and then learn an encoder~$\phi(\cdot)$
from goal~$\bm g$ to design~$\bm \theta$, evaluating the design quality with the trained decoder.
To build a dataset, we have to randomly sample designs and calculate realizations and goals from them, since the physical realization process $U(\cdot)$ is known, and the reverse direction (\ie, goal to design) is difficult as discussed before and is what we are trying to learn.
Thus we first sample~$D$ designs~$\bm \theta_1, \cdots, \bm \theta_D$ and build our dataset as~${\mathcal{D} \coloneqq \bigl\{(\bm \theta_1, \bm u_1, \bm g_1), \, \cdots, (\bm \theta_D, \bm u_D, \bm g_D)\bigr\}}$,
where~${\bm u_i \coloneqq U(\bm \theta_i)}$ and~$\bm g_i$ is the goal calculated from the realization~$\bm u_i$.
The calculation of the goal~$\bm g_i$ summarizes properties that we care about in~$\bm u_i$, and this process depends on the synthesis problem itself and how the cost function~$\mathcal{L}_{\cdot}(\cdot, \cdot)$ is designed.
We split $\mathcal{D}$ into $\mathcal{D}_{\texttt{train}}$, $\mathcal{D}_{\texttt{val}}$, and $\mathcal{D}_{\texttt{test}}$.
We then train our surrogate $\hat{U}(\cdot)$ such that
\begin{equation}
    \label{eqn:decoder}
    \hat{U}^*(\cdot) \coloneqq \arg \min_{\hat{U}(\cdot)} \mathbb{E}_{(\bm \theta, \bm u, \cdot) \sim \mathcal{D}_{\texttt{train}}} \bigl[ ||\hat{U}(\bm \theta) - \bm u||^2 \bigr],
\end{equation}
so that $\hat{U}^*(\cdot)$ serves as a differentiable surrogate of the physical realization function $U(\cdot)$. We then use the trained decoder $\hat{U}^*(\cdot)$ to train an encoder $\phi(\cdot)$:
\begin{equation}
    \label{eqn:encoder}
    \phi^*(\cdot) \coloneqq \arg \min_{\phi(\cdot)} \mathbb{E}_{(\cdot, \cdot, \bm g) \sim \mathcal{D}_{\texttt{train}}} \bigl[ \mathcal{L}_{\bm g}\bigl(\phi(\bm g), \hat{U}^*(\phi(\bm g))\bigr) \bigr].
\end{equation}
Note that although our method resembles an autoencoder, it is not an autoencoder, since we learn the decoder first and then the encoder rather than jointly. Besides, the design space is usually not a lower-dimensional representation of the goal space, and the cost function is usually not simply reconstruction loss.

\section{Empirical evaluation approach}
\label{sec:eval}

In this work, we use our method to solve two real problems as case studies: a task of optimizing the extruder path in additive manufacturing, and a task of actuating a soft robot in an inverse kinematics setting.
While the details of the two differ, we evaluate them in the same way.
In each case, we compare our algorithm with two baselines---\texttt{direct-learning} and \texttt{direct-optimization}---and evaluate our method in terms of relative quality and run time.

\paragraph{\texttt{direct-learning}:}
One natural baseline is to directly learn a model from the goal $\bm g$ to the design~$\bm \theta$. Thus we introduce the \texttt{direct-learning} model $\phi_{\texttt{dl}}(\cdot)$:
\begin{equation}
    \label{eqn:direct-learning}
    \phi_{\texttt{dl}}(\cdot) \coloneqq \arg \min_{\phi(\cdot)} \mathbb{E}_{(\bm \theta, \cdot, \bm g) \sim \mathcal{D}_{\texttt{train}}} \bigl[ ||\bm \theta - \phi(\bm g)||^2 + \mathcal{R}_{\texttt{dl}}(\phi(\bm g)) \bigr].
\end{equation}
Note that, since there is no surrogate, we do not have access to the realization $\bm u$. We thus have to slightly adjust the cost function into a squared Euclidean distance part and a regularizer part $\mathcal{R}_{\texttt{dl}}(\cdot)$ (if there is one), the latter of which comes from the original cost function $\mathcal{L}_{\cdot}(\cdot, \cdot)$. We expect our method to perform much better than \texttt{direct-learning}.

\paragraph{\texttt{direct-optimization}:}
Since our surrogate is a neural network and therefore differentiable, another natural baseline is to directly optimize the design $\bm \theta$ with respect to the goal $\bm g$ by using a gradient-based optimizer (\eg, BFGS), which provides us a rough performance ``upper bound'' on our method:
\begin{equation}
    \label{eqn:direct-optimization}
    \phi_{\texttt{do}}(\bm g) \coloneqq \arg \min_{\bm \theta} \mathcal{L}_{\bm g}(\bm \theta, \hat{U}^*(\bm \theta)).
\end{equation}
Note that AmorFEA~\citep{xue2020amortized} is state-of-the-art method for PDE-constrained optimization, which uses the same approach as the \texttt{direct-optimization} baseline.
We expect our method to have performance close to \texttt{direct-optimization}, while running orders of magnitude faster.

To evaluate our method and the baselines, we iterate over all $\bm g$ from $\mathcal{D}_{\texttt{test}}$, and evaluate the quality of each tuple $\bigl(\phi(\bm g), U(\phi(\bm g)), \bm g\bigr)$, where we use the physical realization function $U(\cdot)$ rather than the learned surrogate $\hat{U}^*(\cdot)$. We run every experiment 3 times with random initialization of neural networks. More details are provided in each case study and in the appendix.

%% file: text/pt.tex
\section{Case study: extruder path planning}

3D printing (a.k.a.\ additive manufacturing) is the process of creating a 3D object from a 3D model by successively adding layers of material.  It has a wide variety of applications in areas including aerospace, automotive, healthcare, and architecture~\citep{shahrubudin2019overview}. Popular 3D printers use thermoplastic polymers (PLA, ABS, nylon, etc.) as the printing material, but these have limited strength. To address this issue, some recent printers support using strong fibers to reinforce the composite. In this work, we explore a 3D printer (the Markforged Mark Two) capable of extruding discrete fibers (fiberglass, kevlar, or carbon fiber) along a controllable path.  However, since the fibers are stiff and non-stretchable, the printed fiber path will be ``smoothed'' compared to the extruder path. As shown in~\fig{fig:teaser}c, without path planning, the fiber path will be severely deformed. We seek to find a general method that, given any desired fiber path, plans an extruder path to compensate for the deformation caused by the printing process. To the best of our knowledge, there is no existing automated method for this task, so it is worthwhile to tackle it using machine learning.

\subsection{Cost function}
Following the notation in~\sect{sec:method}, we denote the target fiber path as $\bm g \in \mathbb{R}^{n \times 2}$: a path is represented as a series of $n$ points, and $n$ varies from path to path (at the scale of hundreds in our experiments). We denote the extruder path (the design) as~${\bm \theta \in \mathbb{R}^{n \times 2}}$ and its realization fiber path as~${\bm u \in \mathbb{R}^{n \times 2}}$. Our cost function~$\mathcal{L}_{\cdot}(\cdot, \cdot)$ is defined as:
\begin{equation}
    \label{eqn:pt-loss}
    \mathcal{L}_{\bm g}(\bm \theta, \bm u) \coloneqq ||\bm g - \bm u||_2^2 + \lambda \cdot \mathcal{R}(\bm \theta),
\end{equation}
where $\lambda$ is a hyper-parameter and $\mathcal{R}(\cdot)$ is a smoothing regularizer that calculates the sum of squared empirical second-order derivatives of the extruder path:
\begin{equation}
    \label{eqn:pt-reg}
    \small
    \mathcal{R}(\bm \theta) \coloneqq \sum_{i = 2}^{n - 1} \left( \left(\frac{\bm \theta_{i + 1} - \bm \theta_i}{||\bm \theta_{i + 1} - \bm \theta_i||_2} - \frac{\bm \theta_i - \bm \theta_{i - 1}}{||\bm \theta_i - \bm \theta_{i - 1}||_2} \right) \bigg/ \left( \frac{||\bm \theta_{i + 1} - \bm \theta_i||_2 + ||\bm \theta_i - \bm \theta_{i - 1}||_2}{2} \right) \right)^{\! 2}\!,
\end{equation}
where $\bm \theta_i \in \mathbb{R}^2$ is the $i$-th row of $\bm \theta$.

\subsection{Evaluation metric}
\label{sec:pt-metric}

The most intuitive way to evaluate the quality of extruder path~$\bm \theta$ is to measure the distance between~$\bm g$, the desired fiber path, and~$\bm u$, the fiber path we get by printing~$\bm \theta$.
Note that it is likely the model under-estimates or over-estimates the deformation of fiber introduced by printing, so both cases can happen when we print with path $\bm \theta$: we run out of fiber before we finish $\bm \theta$, or there is still some fiber remaining in the nozzle after we finish $\bm \theta$ (the total length of fiber is fixed given a desired fiber path $\bm g$). In other words, the $\bm g_i$'s and $\bm u_i$'s might not be synchronized. Thus, directly measuring the distance between $\bm g_i$ and $\bm u_i$ does not necessarily reflect the difference between desired fiber path and the fiber path we get.
Therefore, to better measure the distance between $\bm g$ and $\bm u$ during testing, we use Chamfer distance~\citep{fan2017point}, which was first proposed by~\citet{barrow1977parametric} as an image matching technique, later developed as a commonly used (semi)metric to measure the difference between two sets, and has been shown to have a higher correlation with human judgment compared to intersection over union and earth mover's distance~\citep{pix3d}:
\begin{equation}
    d_{\texttt{CD}}(\bm g, \bm u) \coloneqq \frac{1}{2} \biggl( \frac{1}{n} \sum_{i = 1}^n \min_{j \in [1, n]} \norm{\bm g_i - \bm u_j}_2 + \frac{1}{n}\sum_{j = 1}^n \min_{i \in [1, n]} \norm{\bm g_i - \bm u_j}_2 \biggr).
\end{equation}

\subsection{Implementation}
\label{sec:pt-implementation}

\input{figText/pt-pipeline}

The pipeline is shown in~\fig{fig:pt-pipeline}. To generate the dataset for calibrating the decoder, we first use elliptical slice sampling~\citep{murray2010elliptical} (New BSD License) to sample random extruder paths from a Gaussian process.  We then use a physical simulator built using Bullet~\citep{Coumans2010Bullet} (zlib License), calibrated to a real printer, to predict the realization (fiber path) for each extruder path.
We generate 10,000 paths, split into 90\% training, 5\% validation, and 5\% testing.
For decoder, encoder, and \texttt{direct-learning}, we use an MLP with 5 hidden layers and ReLU as the activation function.
The MLP takes 61 points as input and produces 1 point as output, and is applied in a ``sliding window'' fashion over the entire path (details in appendix).
We train every model with a learning rate of $1\times10^{-3}$ for 10 epochs using PyTorch~\citep{paszke2019pytorch} and Adam optimizer~\citep{Kingma2015Adam:}.
For \texttt{direct-optimization}, we use the BFGS implementation in SciPy~\citep{2020SciPyNMeth}.
More implementation details are included in the appendix.

\subsection{Experiments}
\label{sec:pt-experiments}
\input{figText/pt-cb-cd}
\input{figText/pt-eval_sim}
\input{figText/pt-runtime}
\input{figText/pt-eval_real}
\paragraph{Fiber path quality evaluated in simulation.}
To quantitatively evaluate the effectiveness of our approach, the most straightforward way is to run its prediction on the simulator and see how close the simulated fiber path is to the input fiber path.
We compare the performance of our approach (encoder), \texttt{direct-learning}, and \texttt{direct-optimization} on the test set of 500 paths, for different values of the regularization parameter $\lambda$.  We report the average Chamfer distance (\ssect{sec:pt-metric}) with the standard error among 3 runs in~\tbl{tbl:pt-cb-cd}.
Note that \texttt{direct-optimization} runs very slowly, so we instead tune its regularizer weight on the first 40 test samples, select the best regularizer weight, and report its performance on the whole test set using the selected regularizer weight (more details in the appendix).
The results demonstrate that our method significantly outperforms \texttt{direct-learning}, and the performance is comparable to \texttt{direct-optimization}.

As a qualitative evaluation, \fig{fig:pt-eval_sim} shows the predictions of both \texttt{direct-learning} and our method on samples from the test set.
We select the run with minimum average Chamfer distance over the test set for both \texttt{direct-learning} and our method, respectively.
The results indicate that our method is better than \texttt{direct-learning} on handling details in the fiber path.
We also visualize the Chamfer distance from each individual point on one path to the other, and observe that the distances for our method are generally lower than for \texttt{direct-learning}.

\paragraph{Running time comparison.}
To train needed neural networks, \texttt{direct-optimization} takes roughly 10 minutes, \texttt{direct-learning} takes roughly 1 hour, and our method takes roughly 5 days. Note that training costs are amortized, since we only need to train once.
We then evaluate the inference time of the three algorithms on a server with two Intel(R) Xeon(R) E5-2699 v3 CPUs running at 2.30GHz. Since small neural networks generally run faster on the CPU, we run all of the tests solely on CPU. We run every algorithm on the first 50 paths in the test set and report the average inference time in~\tbl{tbl:pt-runtime}. As we can see, both ours and \texttt{direct-learning} achieve a running time below 1 millisecond, while \texttt{direct-optimization} runs orders of magnitude slower.

\paragraph{Fiber path quality evaluated on a real printer.}
Lastly, we test our extruder path solutions on a real Markforged Mark Two 3D printer. We set the desired fiber path to a star, and print the star itself (without planning) as well as solutions from both \texttt{direct-learning} and our method. We select regularizer weights based on~\tbl{tbl:pt-cb-cd}, \ie, 0.1 for \texttt{direct-learning}, 1.5 for our method, and we show two of the trained models. The results are visualized in~\fig{fig:pt-eval_real}, confirming that our method successfully improves the quality of the printed fiber.

%% file: figText/pt-pipeline.tex
\begin{figure*}[t]
    \centering
    \includegraphics[width=\linewidth]{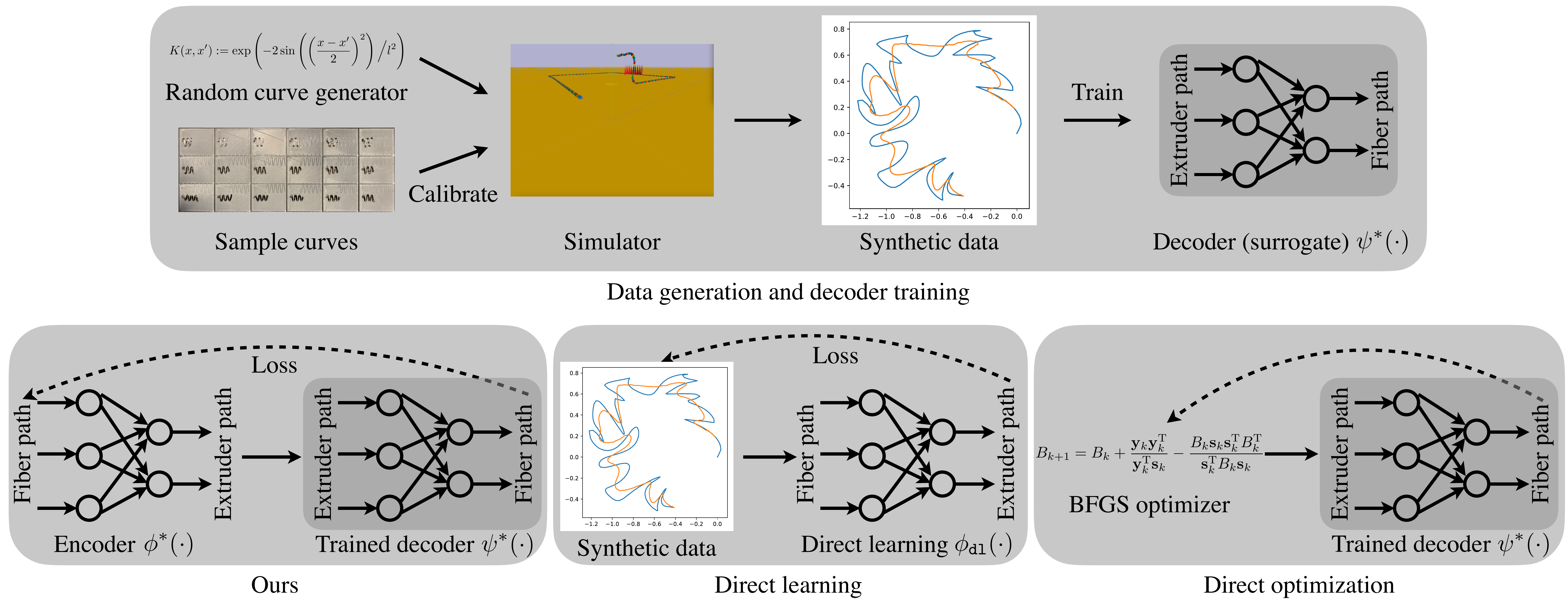}
    \caption{The pipeline for our method and two baselines for extruder path planning. We first generate data by building a simulator and calibrating it using a real printer, and we train the decoder and \texttt{direct-learning} using the synthetic dataset. We then train our method and build \texttt{direct-optimization} using the trained decoder.}
    \label{fig:pt-pipeline}
\end{figure*}

%% file: figText/pt-cb-cd.tex
\begin{table}[t]
    \small
    \setlength{\tabcolsep}{2pt}
    \caption{Path-planning evaluation of the average Chamfer distance on the test set evaluated in simulation}
    \vskip 0.1in
 	\centering
    \begin{tabular}{cccccc}
    \toprule
    Regularizer weight & 0.1 & 0.3 & 0.6 & 1.0 & 1.5 \\
    \midrule
    \texttt{direct-learning} & 0.0314$\pm$0.0008 & 0.0319$\pm$0.0016 & 0.0502$\pm$0.0072 & 0.1007$\pm$0.0292 & 0.1457$\pm$0.0216 \\
    Ours & \textbf{0.0180$\pm$0.0004} & \textbf{0.0157$\pm$0.0003} & \textbf{0.0164$\pm$0.0004} & \textbf{0.0158$\pm$0.0002} & \textbf{0.0156$\pm$0.0002} \\
    \midrule
    \texttt{direct-optimization} & \multicolumn{5}{c}{0.0155$\pm$0.0002} \\
    \bottomrule
    \end{tabular}
    \label{tbl:pt-cb-cd}
\end{table}

%% file: figText/pt-eval_sim.tex
\begin{figure*}[t]
    \centering
    \includegraphics[width=\linewidth]{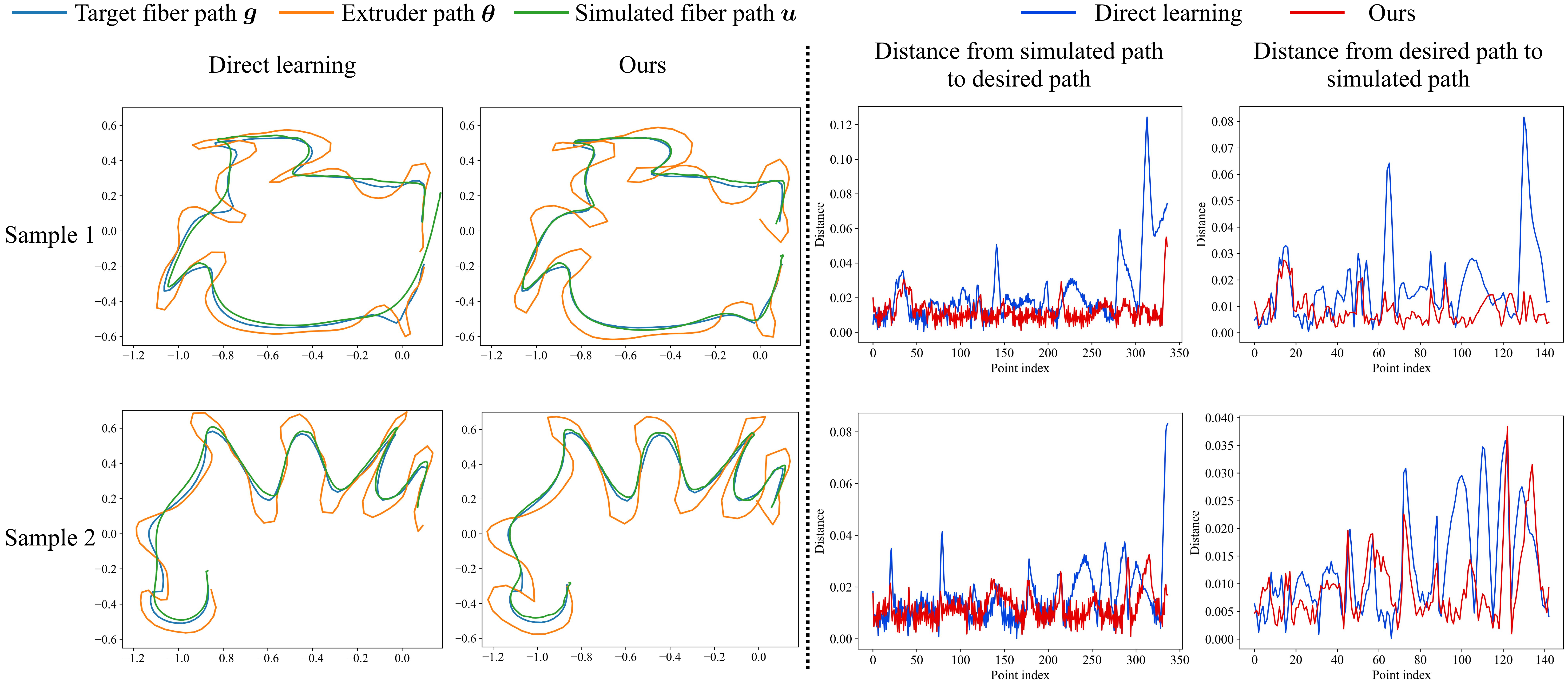}
    \caption{Path-planning evaluation of \texttt{direct-learning} \textit{vs.} ours on the test set evaluated in simulation. We also visualize the Chamfer distance for each point on both the simulated fiber path and the desired fiber path.}
    \label{fig:pt-eval_sim}
\end{figure*}

%% file: figText/pt-runtime.tex
\begin{wrapfigure}{R}{0.45\textwidth}
    \captionof{table}{Path-planning evaluation of the average running time on the first 50 samples in the test set}
 	\centering
    \begin{tabular}{rl}
    \toprule
     & Avg. time (s) \\
    \midrule
    \texttt{direct-learning} & 7.96$\times$10$^{-\text{4}}$ \\
    Ours & 7.96$\times$10$^{-\text{4}}$ \\
    \texttt{direct-optimization} & 1.17$\times$10$^{\text{4}}$ \\
    \bottomrule
    \end{tabular}
    \label{tbl:pt-runtime}
\end{wrapfigure}

%% file: figText/pt-eval_real.tex
\begin{figure*}[t]
    \centering
    \includegraphics[width=0.8\linewidth]{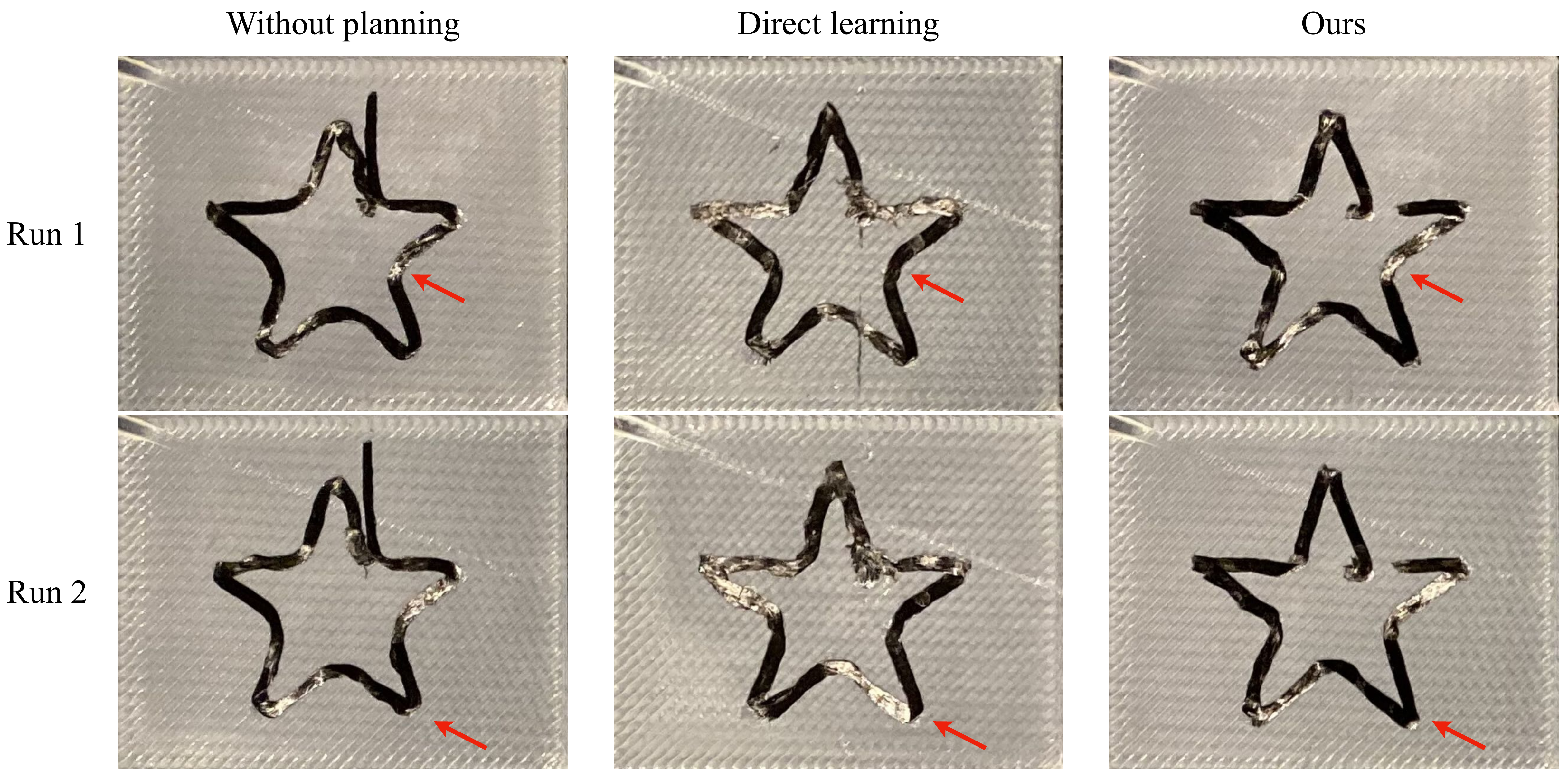}
    \caption{Path-planning evaluation of ``without planning'' \textit{vs.} \texttt{direct-learning} \textit{vs.} ours on Markforged Mark Two, with a star as the desired fiber. For ``without planning'', we have the same extruder path for the 2 runs; for \texttt{direct-learning} and ours, the 2 runs are predictions from neural networks trained with different initializations.}
    \label{fig:pt-eval_real}
\end{figure*}

%% file: text/rb.tex
\section{Case study: constrained soft robot inverse kinematics}

Partial differential equations (PDEs) are a powerful tool for describing complex relationships between variables and have been used widely in areas including physics, engineering, and finance.
In PDE-constrained optimization~\citep{biegler2003large}, the goal is to optimize a cost function such that the constraints can be written as PDEs, \ie, the solutions are consistent with the relationships specified by the PDE.
In the context of synthesis problems, we can consider the boundary conditions of the PDE as the design, and the solution of the PDE as the realization.
Similar to other synthesis problems, different boundary conditions can result in the same PDE solution, and the cost function may not have a unique minimum.
In the above situations, we propose to apply our method. We test our method on a specific PDE-constrained optimization problem---constrained soft robot inverse kinematics, which serves as a representative use case of our method in this large category of problems. 

Soft robots made of elastic materials, have received significant recent attention because of their reduced potential harm when working with humans~\citep{rus2015design}. Researchers have explored a variety of applications, including surgical assistance~\citep{cianchetti2014soft}, bio-mimicry~\citep{li2021self}, and human-robot interaction~\citep{pang2020coboskin}.
In this case study, as in~\citet{xue2020amortized}, we use a snake-like soft robot with a fixed bottom, in which we can control the stretch ratios on both sides of the robot. As shown in~\fig{fig:teaser}d, the objective is for the midpoint at the top of the robot to reach a target, while making sure that the robot does not collide with a fixed-size circular obstacle. The relationship between the robot's shape and the stretch ratios can be written as a PDE, and as shown in~\fig{fig:teaser}d, there are different solutions to achieve the goal.

\subsection{Cost Function}
We adopt the soft robot from~\citet{xue2020amortized}, which has an original height of 10 and an original width of 0.5, with its bottom is fixed. The goal vector $\bm g$ is in $\mathbb{R}^{2 \times 2}$, where $\bm g_1 \in \mathbb{R}^2$ indicates the target location and $\bm g_2 \in \mathbb{R}^2$ indicates the obstacle location. We denote the radius of the obstacle as $r$, which we set to $0.9$.
The design (control) vector $\bm \theta$ is in $\mathbb{R}^{n}$ with $n=40$, with $\theta_i \in \mathbb{R}$ indicating the stretch ratio of the $i$-th segment (\eg, $\theta_i = 0.95$ indicates the $i$-th segment is contracted by 5\%).
The physical realization vector $\bm u$ is in $\mathbb{R}^{m \times 2}$ with $m=103$, where $\bm u_i \in \mathbb{R}^2$ indicates the location of the $i$-th vertex on the soft robot's mesh. The location of the top midpoint is denoted as $\bm u_{\texttt{tm}} \in \mathbb{R}^2$. 
Implicitly, the relationship between $\bm \theta$ and $\bm u$ obeys a PDE that governs the deformation of the robot, as detailed in~\citet{xue2020amortized}.

The cost function $\mathcal{L}_{\cdot}(\cdot, \cdot)$ is defined as
\begin{equation}
    \label{eqn:rb-obj}
    \mathcal{L}_{\bm g}(\bm \theta, \bm u) \coloneqq \frac{1}{2} ||\bm g_1 - \bm u_{\texttt{tm}}||_2^2 + \lambda_1 \cdot \mathcal{B}(\bm u, \bm g_2) + \lambda_2 \cdot \mathcal{R}(\bm \theta).
\end{equation}
The first term $||\bm g_1 - \bm u_{\texttt{tm}}||_2^2$ is the squared Euclidean distance between the top midpoint of the robot and the target. The second term, weighted by a hyper-parameter $\lambda_1$ (which we fix at 0.5), enforces the constraint via a barrier function \citep{nesterov2018lectures, nocedal2006numerical} for the obstacle $\mathcal{B}(\bm u, \bm g_2)$:
\begin{equation}
    \label{eqn:rb-barr}
    \mathcal{B}(\bm u, \bm g_2) \coloneqq \frac{1}{m} \sum_{i = 1}^m \big(\max(r + \Delta r - ||\bm u_i - \bm g_2||_2, 0) \big)^2,
\end{equation}
where $\Delta r$ is a hyper-parameter (which we fix at 0.1). A positive $\Delta r$ provides a penalty as well as nonzero gradients when the robot gets close to the obstacle.
The last term contains another hyper-parameter $\lambda_2$, varied in our experiments, weighting a smooth regularization term $\mathcal{R}(\bm \theta)$, with
\begin{equation}
    \label{eqn:rb-reg}
    \mathcal{R}(\bm \theta) \coloneqq \frac{1}{n - 4} \sum_{\substack{1 < i < n, i \neq n/2, \\ i \neq n/2 + 1}} \left( \frac{\theta_{i + 1} - \theta_i}{2} - \frac{\theta_i - \theta_{i - 1}}{2} \right)^{\! 2}\!,
\end{equation}
where~$\theta_i$ for~${i = 1, 2, \cdots, n/2}$ corresponds to stretch ratios on the left-hand side of the robot, and~$\theta_i$ for~${i = n/2+1, \cdots, n}$ corresponds to stretch ratios on the right-hand side of the robot.
This regularizer prevents unphysical deformations with strong discontinuities. 

\subsection{Evaluation metric}

Since there are two objectives---``reach'' and ``avoid''---in this task, we have two evaluation metrics. The first metric is the number of cases that successfully avoid the obstacle (\ie, have all vertex positions outside the obstacle circle). The second metric is the average Euclidean distance of the robot's top midpoint to the target for successful cases.

\subsection{Implementation}
\label{sec:rb-implementation}
Using the finite element method~\citep{hughes2012finite} and the code from~\citet{xue2020amortized} (MIT license), we randomly generate 40,000 data samples, and split them into 90\% training, 7.5\% validation, 2.5\% testing. For encoder, decoder, and \texttt{direct-learning}, we use an MLP with 3 hidden layers and ReLU activation. We train every model for 200 epochs with a learning rate of $1\times10^{-3}$ using PyTorch~\citep{paszke2019pytorch} and Adam optimizer~\citep{Kingma2015Adam:}. For \texttt{direct-optimization}, we use the BFGS implementation in SciPy~\citep{2020SciPyNMeth}. More implementation details are included in the appendix.

\subsection{Experiments}
\label{sec:rb-experiments}

\input{figText/rb-avoid-obj}
\input{figText/rb-eval_avoid}

\paragraph{Design quality evaluation.}
We experiment with different regularizer weights $\lambda_2$ for our method and the two baselines. During training, we randomly sample the location of the obstacle, and we ensure the robot never collides with the obstacle for \texttt{direct-learning}, since it does not have access to the realization vector and thus its loss function cannot contain the barrier function term for the obstacle (more details about \texttt{direct-learning} are in the appendix).
For a fair comparison, during testing, we set the random seed to 0 such that for the same test sample, the obstacle will appear at the same location for all algorithms.
The number of cases in which the robot successfully avoids the obstacle, with standard error for 3 runs, is shown in~\tbl{tbl:rb-avoid-succ}.  The average Euclidean distance to the target for successful cases, with its standard error, is shown in~\tbl{tbl:rb-avoid-dis}. As the numbers show, our method is competitive to \texttt{direct-optimization}, and performs much better than \texttt{direct-learning}. Samples from the test set are shown in~\fig{fig:rb-eval_avoid} (for both algorithms, we select the best run with a regularizer weight of 0.5). We can see that our method collides less frequently while reaching the target more accurately than \texttt{direct-learning}.

\input{figText/rb-runtime}

\paragraph{Running time.}
To train the neural networks, \texttt{direct-optimization} takes roughly 2 hours, \texttt{direct-learning} takes roughly 4 hours, and our method takes roughly 4.5 hours. Note that training costs are amortized, since we only need to train once.
We then test all algorithms on a server with two Intel(R) Xeon(R) E5-2699 v3 CPUs running at 2.30GHz.  Everything runs solely on the CPU, maximizing efficiency for the small neural networks we use. The results are shown in~\tbl{tbl:rb-runtime}, showing that we successfully reduce the running time from over 1 second to less than 1 millisecond. Note that the soft robot is relatively small (40 control variables), and the time complexity of BFGS grows quadratically w.r.t.\ the number of parameters. Therefore, for more complex soft robots or PDE-constrained optimization problems with a larger number of variables, the running time advantage of our method can be of even greater importance.

%% file: figText/rb-avoid-obj.tex
\begin{table}[t]
    \setlength{\tabcolsep}{3pt}
    \caption{Soft-robot evaluation of the number of successful cases (over 1,000) on test set}
    \vskip 0.1in
 	\centering
    \begin{tabular}{ccccc}
    \toprule
    Regularizer weight & 0.03 & 0.05 & 0.07 & 0.09 \\
    \midrule
    \texttt{direct-learning} & 907.7$\pm$3.1 & 918.3$\pm$3.4 & 910.7$\pm$3.4 & 912.3$\pm$3.8 \\
    Ours & \textbf{986.3$\pm$0.5} & \textbf{975.0$\pm$3.9} & \textbf{981.7$\pm$5.0} & \textbf{984.7$\pm$4.5} \\
    \midrule
    \texttt{direct-optimization} & 997.0$\pm$0.5 & 998.0$\pm$0.0 & 998.3$\pm$0.7 & 997.7$\pm$0.5 \\
    \bottomrule
    \end{tabular}
    \label{tbl:rb-avoid-succ}
\end{table}

\begin{table}[t]
    \setlength{\tabcolsep}{3pt}
    \caption{Soft-robot evaluation of the average distance to the target on successful cases on test set}
    \vskip 0.1in
 	\centering
    \begin{tabular}{ccccc}
    \toprule
    Regularizer weight & 0.03 & 0.05 & 0.07 & 0.09 \\
    \midrule
    \texttt{direct-learning} & 0.2171$\pm$0.0016 & 0.2200$\pm$0.0028 & 0.2236$\pm$0.0006 & 0.2179$\pm$0.0036 \\
    Ours & \textbf{0.0657$\pm$0.0093} & \textbf{0.0464$\pm$0.0018} & \textbf{0.0599$\pm$0.0097} & \textbf{0.0691$\pm$0.0224} \\
    \midrule
    \texttt{direct-optimization} & 0.0233$\pm$0.0003 & 0.0240$\pm$0.0004 & 0.0241$\pm$0.0004 & 0.0242$\pm$0.0003 \\
    \bottomrule
    \end{tabular}
    \label{tbl:rb-avoid-dis}
\end{table}

%% file: figText/rb-eval_avoid.tex
\begin{figure*}[t]
    \centering
    \includegraphics[width=0.7\linewidth]{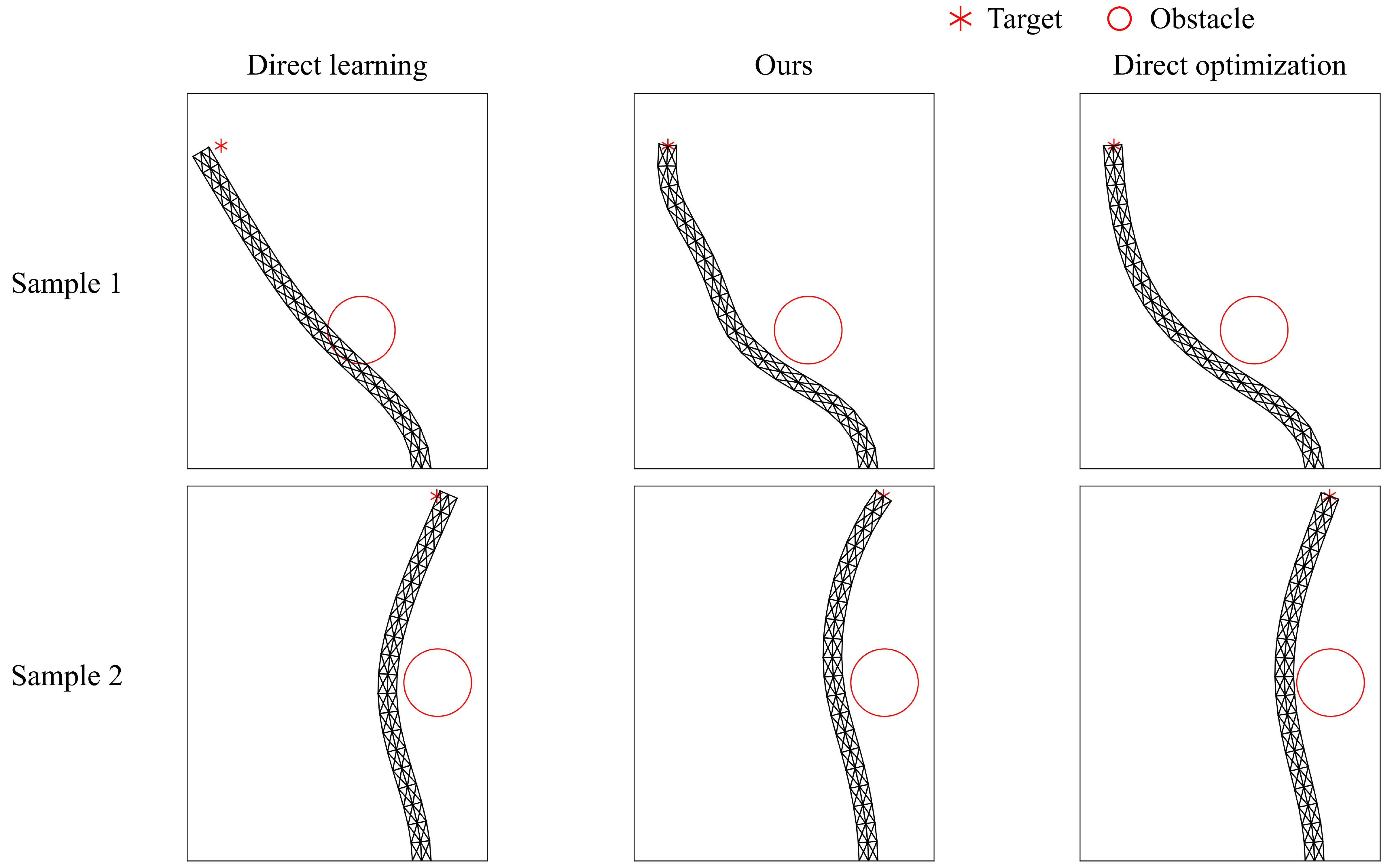}
    \caption{Soft-robot evaluation of \texttt{direct-learning} \textit{vs.} ours \textit{vs.} \texttt{direct-optimization} on examples from the test set.  The direct learning baseline both violates the constraints (Sample 1) and fails to reach the target (Samples 1 and 2), while the run time of \texttt{direct-optimization} is 4000 times that of our method.}
    \label{fig:rb-eval_avoid}
\end{figure*}

%% file: figText/rb-runtime.tex
\begin{wrapfigure}{R}{0.45\textwidth}
    \captionof{table}{Soft-robot evaluation of the average running time on the test set}
 	\centering
    \begin{tabular}{rl}
    \toprule
     & Avg. time (s) \\
    \midrule
    \texttt{direct-learning} & 3.12$\times$10$^{-\text{4}}$ \\
    Ours & 3.34$\times$10$^{-\text{4}}$ \\
    \texttt{direct-optimization} & 1.33$\times$10$^{\text{0}}$ \\
    \bottomrule
    \end{tabular}
    \label{tbl:rb-runtime}
\end{wrapfigure}

%% file: text/discussion.tex
\section{Discussion}
\label{sec:discussion}

In this work, we provided an amortized approach to synthesis problems in machine learning. To tackle the non-differentiability of physical system realizations, the huge computational cost of realization processes, and the non-uniqueness of the design solution, we designed a two-stage neural network architecture, where we first learn the decoder, a surrogate that approximates the realization processes, and then learn the encoder, which proposes a design for an input goal. We tested our approach on two case studies on fiber extruder path planning and constrained soft robot inverse kinematics, where we demonstrated that our method provides designs with much higher quality than supervised learning of the design problem, while being competitive in quality to and orders of magnitude faster than direct optimization of the design solution.

Although the experiments in both case studies show the effectiveness of our approach, we would like to mention some limitations of our method.
First, to effectively learn a differentiable surrogate for the realization process, we need to be able to generate a substantial number of viable designs.  We also need a simulator to calculate physical realizations of them, and the realization process has to be deterministic, although extensions might consider probability distributions over realizations.
Also, to train the encoder, we need the objective (``goal'') to be quantifiable.
Our method provides the greatest gains if the realization is computationally expensive and/or non-differentiable, or if our encoder can exploit the non-uniqueness of designs to choose one good option where supervised learning would have learned a poor ``average'' solution.
Additionally, due to the cost of neural network training, amortization is only a good idea when we need to solve one design problem many times with different goals, or we need fast inference.
From a societal point of view, the primary negative consequence is the potential for replacing human labor in design.
We view the present approach, however, as part of larger human-in-the-loop design processes in line with other software tools for modeling and fabrication.

%% file: text/acknowledgements.tex
\section*{Acknowledgements}

We would like to thank Geoffrey Roeder for helping set up the 3D printer, Amit Bermano, Jimmy Wu, and members of the Princeton Laboratory for Intelligent Probabilistic Systems for valuable discussions and feedback, as well as Markforged.
This work is partially supported by the Princeton School of Engineering and Applied Science, as well as the U.\,S.~National Science Foundation under grants \#IIS-1815070 and \#IIS-2007278.

%% file: text/appendix-pt.tex
\section{Details about extruder path planning}

\subsection{Methods}
\label{sec:pt-methods}

\paragraph{Ours.}
As we discussed before in~\sect{sec:method}, we train a decoder (surrogate) using~\eqn{eqn:decoder} and an encoder using~\eqn{eqn:encoder}, with the cost function defined in~\eqn{eqn:pt-loss}. We take the trained encoder $\phi^*(\cdot)$ as our final model in use.

\paragraph{\texttt{direct-learning}.}
As we discussed in~\sect{sec:eval}, we train \texttt{direct-learning} using~\eqn{eqn:direct-learning}, with a regularizer as defined in~\eqn{eqn:pt-reg}. Note that this is equivalent to training to minimize the cost function $\mathcal{L}_{\cdot}(\cdot, \cdot)$ in~\eqn{eqn:pt-loss}.

\paragraph{\texttt{direct-optimization}.}
As described in~\sect{sec:eval}, we build \texttt{direct-optimization} as in~\eqn{eqn:direct-optimization}, with a trained surrogate of the physical realization process. Here, to enforce that points $\bm \theta_i$ are evenly distributed along the extruder path, we use a slightly different cost function $\mathcal{L}_{\texttt{do}, \cdot}(\cdot, \cdot)$:
\begin{equation}
    \label{eqn:pt-opt-loss}
    \mathcal{L}_{\texttt{do}, \bm g}(\bm \theta, \bm u) \coloneqq d_{\texttt{do}}(\bm g, \bm u) + \lambda_{\texttt{do}} \cdot \mathcal{R}_{\texttt{do}}(\bm \theta),
\end{equation}
where we have a distance function $d_{\texttt{do}}(\cdot, \cdot)$ and a smooth regularizer $\mathcal{R}_{\texttt{do}}(\cdot)$.
The smooth regularizer $\mathcal{R}_{\texttt{do}}(\cdot)$ is derived from~\eqn{eqn:pt-reg} by requiring $||\bm \theta_{i + 1} - \bm \theta_i||_2$ to be the same for all $i$:
\begin{equation}
    \mathcal{R}_{\texttt{do}}(\bm \theta) \coloneqq \frac{1}{S_{\bm \theta}}\sum_{i = 2}^{n - 1} \left( \frac{\bm \theta_{i + 1} - \bm \theta_i}{2} - \frac{\bm \theta_i - \bm \theta_{i - 1}}{2} \right)^2,
\end{equation}
where $S_{\bm \theta}$ is the length of the extruder path $\bm \theta$;  this intrinsically enforces points in $\bm \theta$ to be evenly spaced.
To measure the distance between $\bm g$ and $\bm u$, we first map them into two functions $\bm f_{\bm g}(\cdot)$ and $\bm f_{\bm u}(\cdot)$, such that $\bm f_{\bm g}(s) \in \mathbb{R}^2$ is the location if we walk a distance $s$ along the path $\bm g$ (assuming the path is piecewise linear), and similarly for $\bm f_{\bm u}(\cdot)$. Then the distance function is defined as
\begin{equation}
    d_{\texttt{do}}(\bm g, \bm u) \coloneqq \int_0^1 \bigl\|\bm f_{\bm g}(x \cdot S_{\bm g}) - \bm f_{\bm u}(x \cdot S_{\bm u})\bigr\|^2 \, \mathrm{d} x,
\end{equation}
where $S_{\bm g}$ and $S_{\bm u}$ denote the lengths of paths $\bm g$ and $\bm u$, respectively.

\subsection{Data generation}
\label{sec:pt-data-gen}

\paragraph{Random curve generation.}
To build the dataset, we first need to generate some random 2D curves, which can be used as extruder paths later. The curves should be smooth and non-intersecting.
For each path, we take both the $x$ and $y$ to be Gaussian processes whose kernel function $K(\cdot, \cdot)$ is 
\begin{align}
K(x, x') \coloneqq 
\exp \left(-\frac{\sin^2 \bigl((x-x')/2\bigr)}{2\,l^2}\right),
\end{align}
with the two axes independent and $l=0.1$.
We use elliptical slice sampling~\citep{murray2010elliptical} (New BSD License): for each path, we start from 1,000 points on a unit circle, and sample 1,000 times. To avoid intersections, we use a log-likelihood of $-\infty$ for a self-intersecting path, and a log-likelihood of 0 for a non-intersecting path. We generate 10,000 paths using this approach.

\paragraph{Simulator.}
\input{figText/pt-calibration}
\input{figText/pt-sample_data}
Since it is time-consuming to print every extruder path we generate in the last step on a real printer, we build a simulation system by using Bullet~\citep{Coumans2010Bullet} to help us generate fiber paths.  The simulator is also used in our evaluation.
We calibrate the simulator on two materials---carbon fiber and Kevlar, respectively.
To calibrate, we print paths shaped as sine functions with amplitudes ranging from 3.0 cm to 6.5 cm on the Markforged Mark Two printer (\fig{fig:pt-calibration-real}).
We then measure the amplitudes of the printed fiber paths, which will be lower than the amplitudes of the extruder paths because of smoothing, and fit lines to actual amplitude vs.~extruder amplitude (\fig{fig:pt-calibration}a).
After that, we perform a grid search on the parameters of the simulator (such as stiffness and friction), run simulations with the sine functions as extruder paths, and collect data pairs of extruder path amplitudes and simulated fiber path amplitudes.
We end up with calibrated simulators for carbon fiber (\fig{fig:pt-calibration}b) and Kevlar (\fig{fig:pt-calibration}c), by selecting the set of parameters for each having the minimum sum of squared distances between the fitted line from~\fig{fig:pt-calibration}a) and the collected simulation data points.
Finally, we run the tuned simulators on the generated extruder paths.
Though we have conducted experiments with both materials, due to space constraints, the experiments in the paper use data generated from the carbon fiber simulator.
We visualize some extruder paths and simulated carbon fiber paths in~\fig{fig:pt-sample_data}.

\subsection{Mapping from a path to another}
\label{sec:pt-mapping}

We have to design a neural network that can take in an input path ($\mathbb{R}^{n \times 2}$) and output another path ($\mathbb{R}^{n \times 2}$), which can be used for encoder, decoder, and \texttt{direct-learning}.
Both paths are sequences of $n$ 2D coordinates.
For the purpose of illustration, we use decoder as an example here. Now the input is the extruder path $\bm \theta$, and the output is the resulting fiber path $\bm u$.
Remember $\bm \theta_i \in \mathbb{R}^2$ is the $i$-th row of $\bm \theta$.
Due to the intrinsic equivariant property of the problem, one natural idea is to have a neural network that takes in a certain number of points near $\bm \theta_i$ in $\bm \theta$ and outputs the corresponding point in $\bm u$ (\ie, $\bm u_i$), and we iterate over every $i$, as a window sliding over $\bm \theta$. Note that we used the same neural network for all $i$'s.

We thus use a multilayer perceptron (MLP), which takes $2m + 1$ points (we set $m=30$) and outputs one point. We take $\bm \theta_i$ as the starting point and resample $m$ points both forward and backward along the path $\bm \theta$. To be specific, as in~\ssect{sec:pt-methods}, we first map $\bm \theta$ into a function $\bm f_{\bm \theta}(\cdot)$ such that $\bm f_{\bm \theta}(s) \in \mathbb{R}^2$ is the location if we start from $\bm \theta_1$ and walk a length of $s$ on the path. We further set $\bm f_{\bm \theta}(s) \coloneqq \bm f_{\bm \theta}(0)$ for $s < 0$ and $\bm f_{\bm \theta}(s) \coloneqq \bm f_{\bm \theta}(S_{\bm \theta})$ for $s > S_{\bm \theta}$, where $S_{\bm \theta}$ is the length of extruder path $\bm \theta$. We denote the distance of walking from $\bm \theta_1$ to $\bm \theta_i$ as $s_i$, \ie, $\bm f_{\bm \theta}(s_i) = \bm \theta_i$. The input to the MLP is:
\begin{equation}
    [\bm f_{-m}, \bm f_{-m + 1}, \cdots, \bm f_{0}, \cdots, \bm f_{m}]^\intercal,
\end{equation}
where
\begin{equation}
    \bm f_i \coloneqq \bm f_{\bm \theta}(s_i + i \cdot s_0) - \bm f_{\bm \theta}(s_i),
\end{equation}
and $s_0 = 0.03$ is the step size.
Since the problem is intrinsically translation-equivariant, we normalize every $\bm f_i$ by subtracting $\bm f_{\bm \theta}(s_i)$, as shown in the above equation.

\input{figText/pt-mlp}
\input{figText/pt-opt-tune}

\subsection{Hyper-parameters and neural network training}
\label{sec:pt-hp-training}

The architecture of the MLP is shown in~\tbl{tbl:pt-mlp}, and we implement it in PyTorch~\citep{paszke2019pytorch}.
We coarsely tuned the architecture, including the number of hidden layers (from 1 to 6) and the size of each hidden layer. We noticed that the accuracy is not largely affected by the architecture, as long as there is at least one hidden layer.
We split the dataset into 90\% training (9,000 paths), 5\% validation (500 paths), and 5\% testing (500 paths), and we use the Adam optimizer~\citep{Kingma2015Adam:} with a learning rate of 1$\times$10$^{-\text{3}}$, a learning rate exponential decay of 0.95 per epoch, and a batch size of 1 (path).
We train every model---the decoder, the encoder, and \texttt{direct-learning}---for 10 epochs.
We use our internal cluster with 7 servers with 14 Intel(R) Xeon(R) CPUs.
For our method and \texttt{direct-learning}, we train them with different regularizer weights $\lambda$ = 0.1, 0.3, 0.6, 1.0, and 1.5.
For \texttt{direct-optimization}, we use the BFGS implementation in SciPy~\citep{2020SciPyNMeth}, with a gradient tolerance of 1$\times$10$^{-\text{7}}$.
Since its running time is extremely long, we tune its regularizer weight on the first 40 test samples (\tbl{tbl:pt-opt-tune}) and select the one with the best performance.
To train the needed neural networks, it takes approximately 10 minutes for \texttt{direct-optimization}, approximately 1 hour for \texttt{direct-learning}, and approximately 5 days for our method.
Note that we only need to train once so that the training costs are amortized.

%% file: figText/pt-calibration.tex
\begin{figure}[t]
    \centering
    \includegraphics[width=\linewidth]{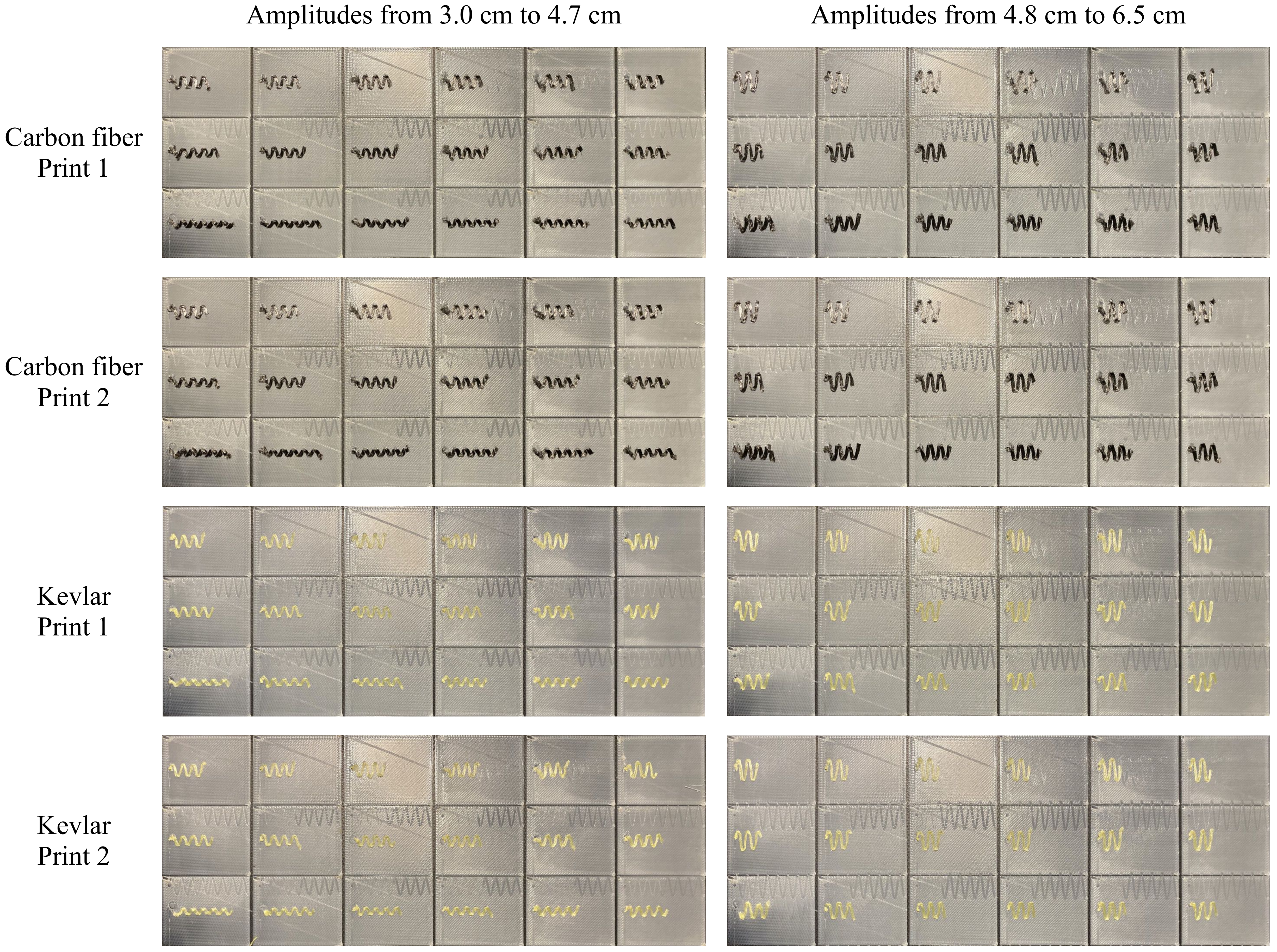}
    \caption{We print paths shaped as sine functions with amplitudes from 3.0 cm to 6.5 cm on the Markforged Mark Two printer, using carbon fiber and Kevlar, respectively. We print everything twice.}
    \label{fig:pt-calibration-real}
\end{figure}

\begin{figure}[t]
    \centering
    \includegraphics[width=\linewidth]{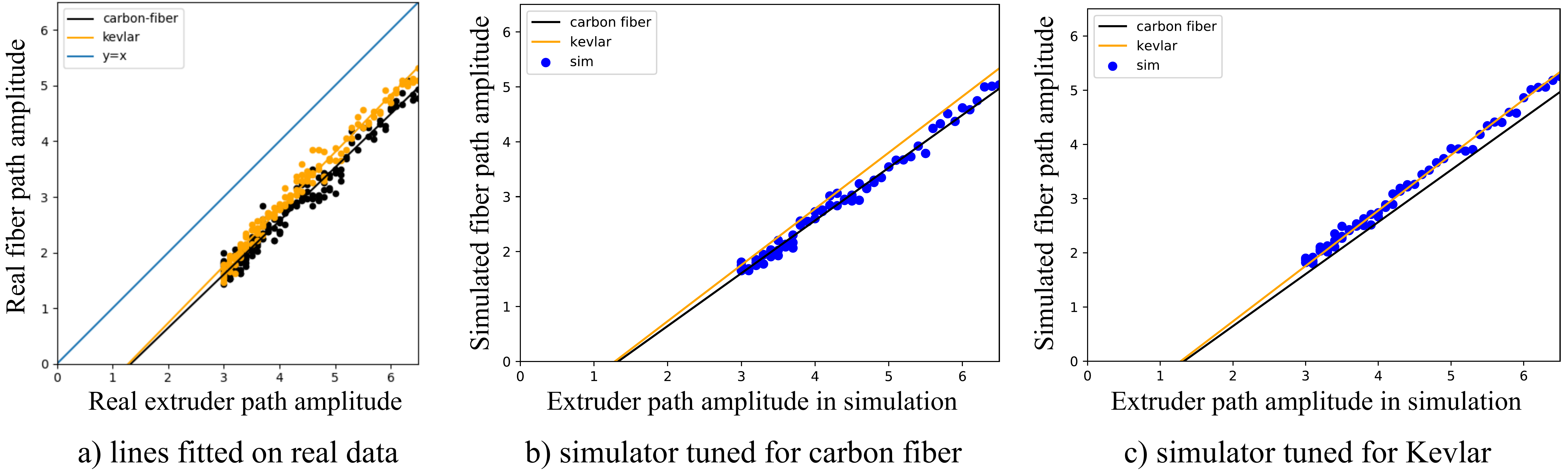}
    \caption{(a) We print paths shaped as sine functions with different amplitudes, collect amplitudes of the printed fiber paths, and fit lines through the data we collected. We experiment with two materials---carbon fiber and Kevlar, and the identity line is also visualized. (b) and (c) We select two sets of simulator hyper-parameters with their simulation results closest to the lines we get from the previous step for carbon fiber and Kevlar, respectively.}
    \label{fig:pt-calibration}
\end{figure}

%% file: figText/pt-sample_data.tex
\begin{figure}[t]
    \centering
    \includegraphics[width=\linewidth]{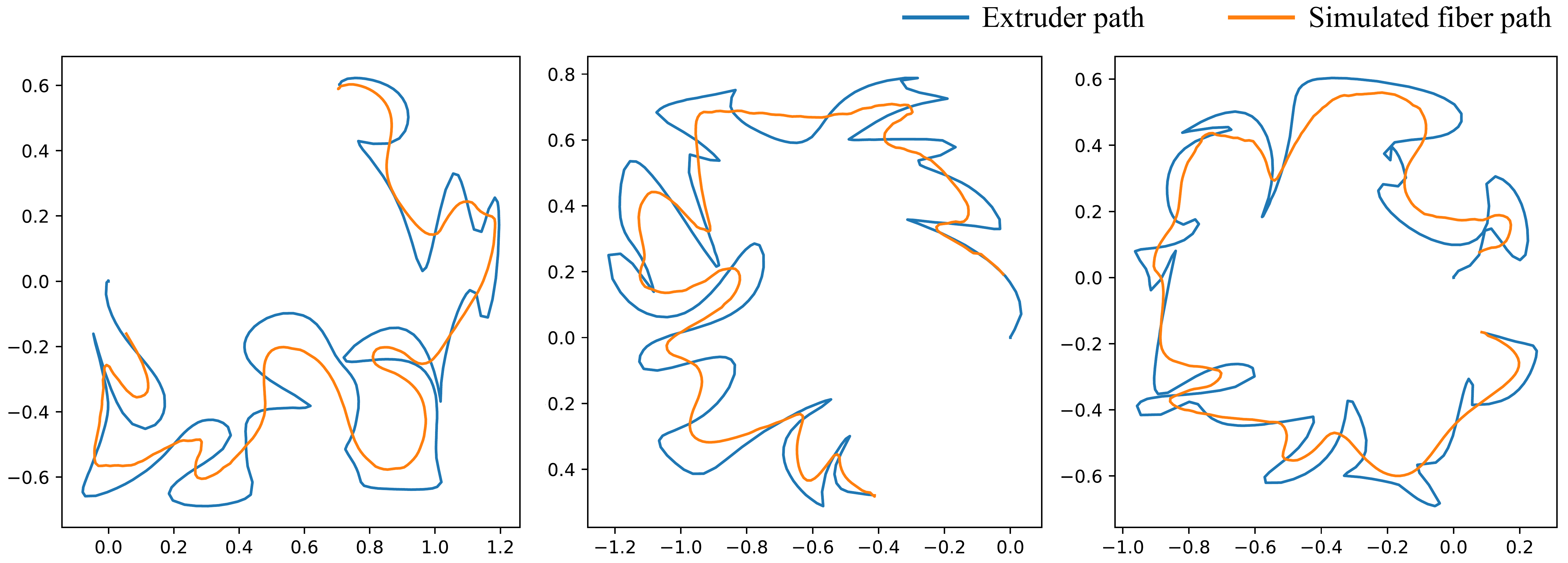}
    \caption{Samples from our dataset. We plot both the extruder paths and the simulated carbon fiber paths.}
    \label{fig:pt-sample_data}
\end{figure}

%% file: figText/pt-mlp.tex
\begin{table}[t]
    \caption{The architecture of the MLP used in extruder path planning}
    \vskip 0.15in
 	\centering
    \begin{tabular}{cc}
    \toprule
    Type & Configurations \\
    \midrule
    Fully connected & 4$m$+2 to 500 \\
    ReLU & N/A \\
    Fully connected & 500 to 200 \\
    ReLU & N/A \\
    Fully connected & 200 to 100 \\
    ReLU & N/A \\
    Fully connected & 100 to 50 \\
    ReLU & N/A \\
    Fully connected & 50 to 25 \\
    ReLU & N/A \\
    Fully connected & 25 to 2 \\
    \bottomrule
    \end{tabular}
    \label{tbl:pt-mlp}
\end{table}

%% file: figText/pt-opt-tune.tex
\begin{table}[t]
    \caption{Path-planning evaluation of \texttt{direct-optimization} of the average Chamfer distance on the first 40 samples in the test set evaluated in simulation}
    \vskip 0.1in
 	\centering
    \begin{tabular}{cccccc}
    \toprule
    Regularizer weight & 0.0001 & 0.0003 & 0.0006 & 0.001 \\
    \midrule
    \texttt{direct-optimization} & 0.0171 & 0.0161 & \textbf{0.0153} & 0.0167 \\
    \bottomrule
    \end{tabular}
    \label{tbl:pt-opt-tune}
\end{table}

%% file: text/appendix-rb.tex
\section{Details about constrained soft robot inverse kinematics}

\input{figText/rb-obs}

\subsection{Robot setting}
We adopt the snake-like soft robot that was used in~\citet{xue2020amortized}.
The robot has an original height of 10 and an original width of 0.5, and its bottom is fixed.
We can control the stretch ratio of 40 segments (20 on the left-hand side and 20 on the right-hand side), as visualized in colors in~\fig{fig:rb-obs}.
The stretch ratios are restricted to be between 0.8 and 1.2.
The physical realization of the robot consists of the locations of its 103 vertexes, as shown in~\fig{fig:rb-obs}.

\subsection{Methods}
\label{sec:rb-methods}

\paragraph{Ours.}
We follow~\sect{sec:method} with the cost function defined as in~\eqn{eqn:rb-obj}, and the obstacle location is randomly sampled.

\paragraph{\texttt{direct-learning}.}
We follow~\sect{sec:eval} with $\mathcal{R}_{\texttt{dl}}(\cdot)$ defined as in~\eqn{eqn:rb-reg}. Note that since we do not have access to the physical realization $\bm u$, we cannot have a barrier function term for the obstacle. Thus, for \texttt{direct-learning}, we still randomly sample the obstacle location, but we guarantee during training, the obstacle does not collide with the robot in every specific training sample.

\paragraph{\texttt{direct-optimization}.}
We follow~\sect{sec:eval} with the cost function defined as in~\eqn{eqn:rb-obj} and the obstacle location randomly sampled. Note that this baseline is similar to the approach used in~\citet{xue2020amortized}. The major difference is that we train the surrogate using supervised loss (as shown in~\eqn{eqn:decoder}), and~\citet{xue2020amortized} trained their surrogate using a physically informed loss that minimizes the total potential energy.

\subsection{Data generation}
We first randomly sample the design vector $\bm \theta$ with each dimension i.i.d.\ uniformly between 0.8 and 1.2. 
For each design vector $\bm \theta$, we solve the governing PDE with the finite element method~\citep{hughes2012finite} to obtain the corresponding physical realization $\bm u$ of the robot.
Note that the obstacle location is randomly sampled during training and randomly sampled with a fixed random seed (we set to 0) during testing.
The center of the obstacle is uniformly sampled from a sector region, as shown in~\fig{fig:rb-obs}, with an angle of 60\degree, an inner radius of 4, and an outer radius of 5.
We altogether generate 40,000 data samples.

\input{figText/rb-mlp}
\input{figText/rb-eval_failure}

\subsection{Hyper-parameters and neural network training}
\label{sec:rb-training}

We use MLP for all models (encoder, decoder, and \texttt{direct-learning}) with ReLU as the activation function and 3 hidden layers of sizes 128, 256, 128, respectively (\tbl{tbl:rb-mlp}).
We coarsely tuned the architecture, including the number of hidden layers (from 1 to 4) and the size of each hidden layer. Similarly, we noticed that the accuracy is not largely affected by the architecture, as long as there is at least one hidden layer.
Note that for the input and output of the neural network, we subtract 1 from all stretch ratios such that they are always between -0.2 and 0.2, and we use displacement of each vertex rather than its absolute location since displacement values are mostly centered around 0. In addition, to ensure that the encoder and \texttt{direct-learning} always output stretch ratios (minus one) between -0.2 and 0.2, we apply a sigmoid layer at the end of both the encoder and \texttt{direct-learning}, and linearly map the sigmoid output to be between -0.2 and 0.2 (as in \tbl{tbl:rb-mlp}). We use the same trick in \texttt{direct-optimization} to ensure the stretch ratios never fall out of range.

We implement all neural networks in PyTorch~\citep{paszke2019pytorch}. We split the dataset into 90\% training (36,000 samples), 7.5\% validation (3,000 samples), and 2.5\% testing (1,000 samples), and we use the Adam optimizer~\citep{Kingma2015Adam:} with a learning rate of~1$\times$10$^{-\text{3}}$, a learning rate exponential decay of 0.98 per epoch, and a batch size of 8. We train every model---the decoder, the encoder, and \texttt{direct-learning}---for 200 epochs.
For our method and all baselines, we experiment with different regularizer weights $\lambda_2$ = 0.03, 0.05, 0.07, and 0.09.
For \texttt{direct-optimization}, we use the BFGS implementation in SciPy~\citep{2020SciPyNMeth}, with a gradient tolerance of 1$\times$10$^{-\text{7}}$.
We use our internal cluster with 7 servers with 14 Intel(R) Xeon(R) CPUs.
To train the needed neural networks, it takes approximately 2 hours for \texttt{direct-optimization}, approximately 4 hours for \texttt{direct-learning}, and approximately 4.5 hours for our method.
Note that since we only need to train once, the training costs are amortized.

\subsection{Failure cases}
Since our encoder and decoder are both neural networks, there might be some generalization errors. We show two failure cases of our method in test set in~\fig{fig:rb-eval_failure}. The design proposed by our method misses the target by a short distance in the first example, and both touches the obstacle and misses the target by a short distance in the second example.

\subsection{Ablation study: linear encoder}
\input{figText/rb-ablation-obj}
\input{figText/rb-ablation}

To demonstrate the non-linearity in our encoder is necessary, we train a linear encoder on the pre-trained non-linear decoder, following exactly the same training procedure mentioned in~\ssect{sec:rb-training}. We show the number of cases the robot successfully avoids the obstacle and the average Euclidean distance to the target for successful cases in~\tbl{tbl:rb-ablation-obj}. Linear encoder violates the obstacle constraints approximately 11.6 times as much as the non-linear encoder, and the average Euclidean distance for successful cases is approximately 2.9 times as much as the non-linear encoder.
Two samples are shown in~\fig{fig:rb-ablation}. In both samples, the linear encoder misses the target by a short distance, and in the second sample, the linear encoder violates the obstacle constraint.

%% file: figText/rb-obs.tex
\begin{figure*}[t]
    \centering
    \includegraphics[width=0.6\linewidth]{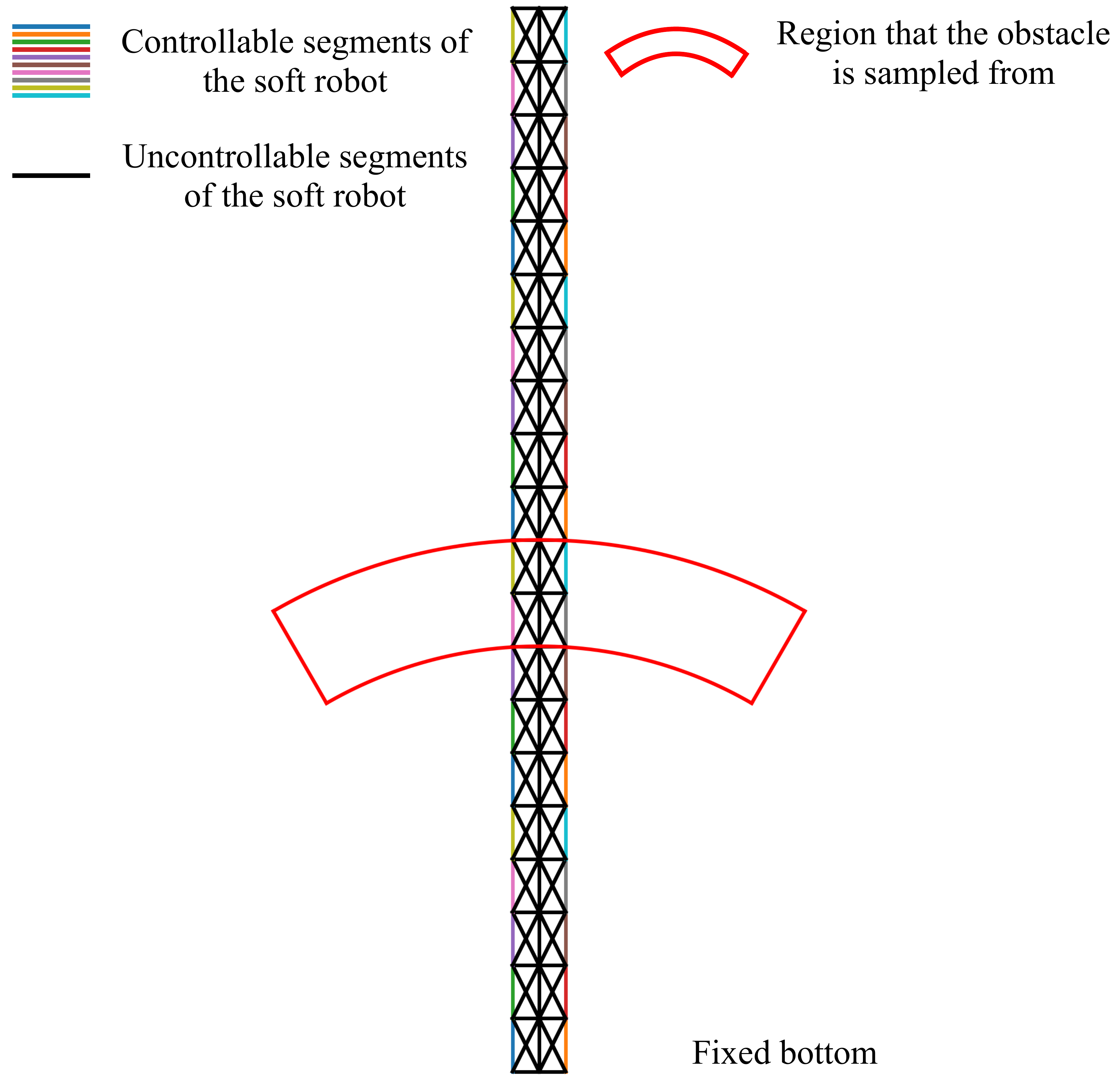}
    \caption{We visualize the soft robot with controllable segments in color, uncontrollable segments in black. The red region shows where we sample the center of the obstacle---a sector region with an angle of 60\degree, inner and outer radiuses of 4 and 5, respectively. }
    \label{fig:rb-obs}
\end{figure*}

%% file: figText/rb-mlp.tex
\begin{table}[t]
    \small
    \setlength{\tabcolsep}{4pt}
    \caption{The architecture of the MLP used in constrained soft robot inverse kinematics}
    \vskip 0.15in
 	\centering
    \begin{tabular}{cccccc}
    \toprule
    \multicolumn{2}{c}{\texttt{direct-learning}} & \multicolumn{2}{c}{Decoder} & \multicolumn{2}{c}{Encoder} \\
    \cmidrule(lr){1-2} \cmidrule(lr){3-4} \cmidrule(lr){5-6}
    Type & Configurations & Type & Configurations & Type & Configurations \\
    \midrule
    Fully connected & 4 to 128 & Fully connected & 40 to 128 & Fully connected & 4 to 128 \\
    ReLU & N/A & ReLU & N/A & ReLU & N/A \\
    Fully connected & 128 to 256 & Fully connected & 128 to 256 & Fully connected & 128 to 256 \\
    ReLU & N/A & ReLU & N/A & ReLU & N/A \\
    Fully connected & 256 to 128 & Fully connected & 256 to 128 & Fully connected & 256 to 128 \\
    ReLU & N/A & ReLU & N/A & ReLU & N/A \\
    Fully connected & 128 to 40 & Fully connected & 128 to 206 & Fully connected & 128 to 40 \\
    Sigmoid & N/A & / & / & Sigmoid & N/A \\
    Linear map & 0.2(2$x-$1) & / & / & Linear map & 0.2(2$x-$1) \\
    \bottomrule
    \end{tabular}
    \label{tbl:rb-mlp}
\end{table}

%% file: figText/rb-eval_failure.tex
\begin{figure*}[t]
    \centering
    \includegraphics[width=\linewidth]{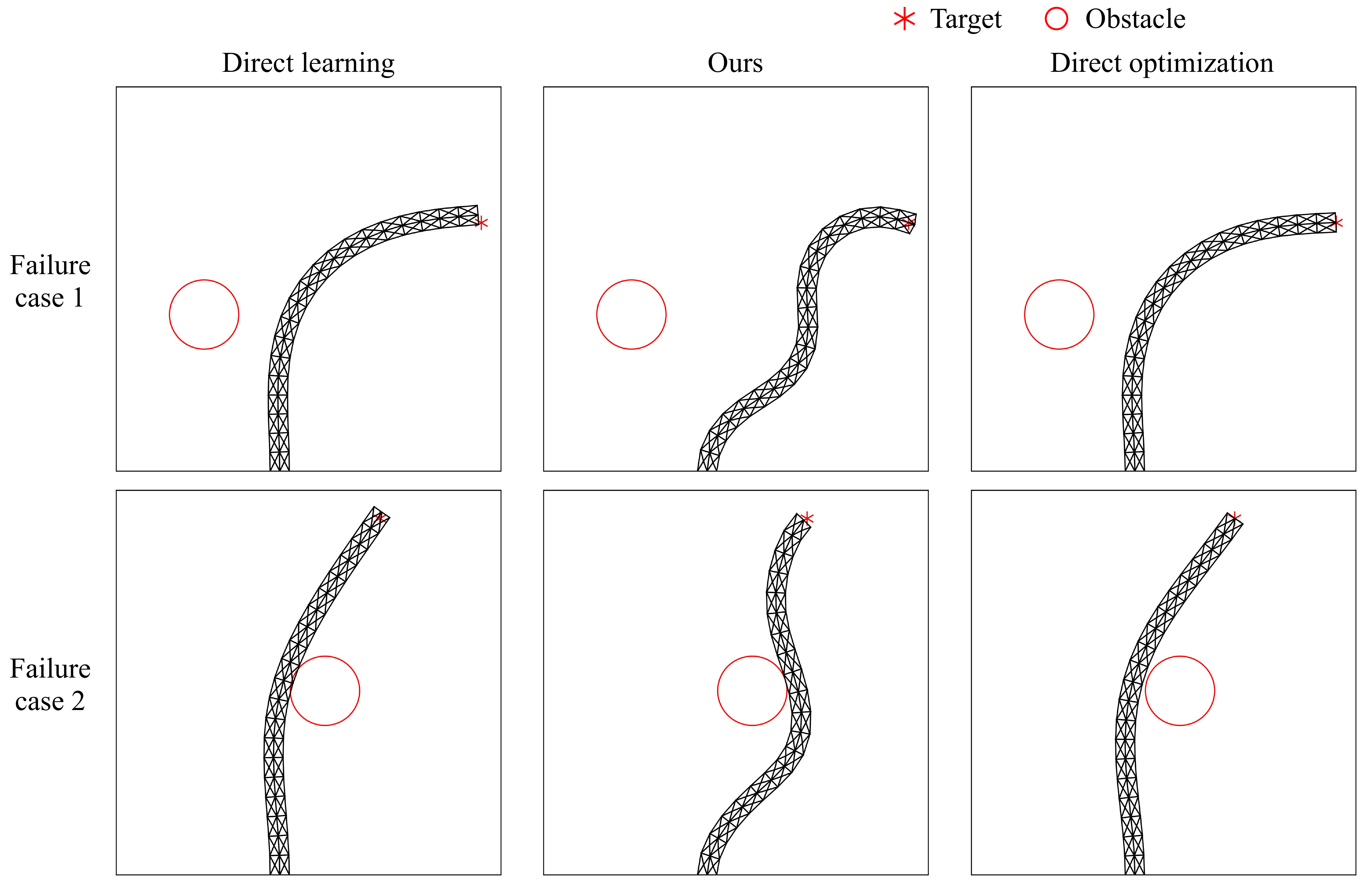}
    \caption{Failure cases of our method in test set on soft robot. In rare cases, our method might miss the target by a short distance (Samples 1 and 2) or touch the obstacle (Sample 2).}
    \label{fig:rb-eval_failure}
\end{figure*}

%% file: figText/rb-ablation-obj.tex
\begin{table}[t]
    \setlength{\tabcolsep}{3pt}
    \caption{Ablation study: linear encoder vs. non-linear encoder for soft-robot evaluated on test set. All encoders are trained on non-linear decoders with a regularizer weight of 0.05}
    \vskip 0.1in
 	\centering
    \begin{tabular}{ccc}
    \toprule
     & \#successful cases (over 1,000) & Avg. distance to target on successful cases \\
    \midrule
    Non-linear encoder & \textbf{975.0$\pm$3.9} & \textbf{0.0464$\pm$0.0018} \\
    Linear encoder & 710.7$\pm$24.8 & 0.1324$\pm$0.0276 \\
    \bottomrule
    \end{tabular}
    \label{tbl:rb-ablation-obj}
\end{table}

%% file: figText/rb-ablation.tex
\begin{figure*}[t]
    \centering
    \includegraphics[width=0.7\linewidth]{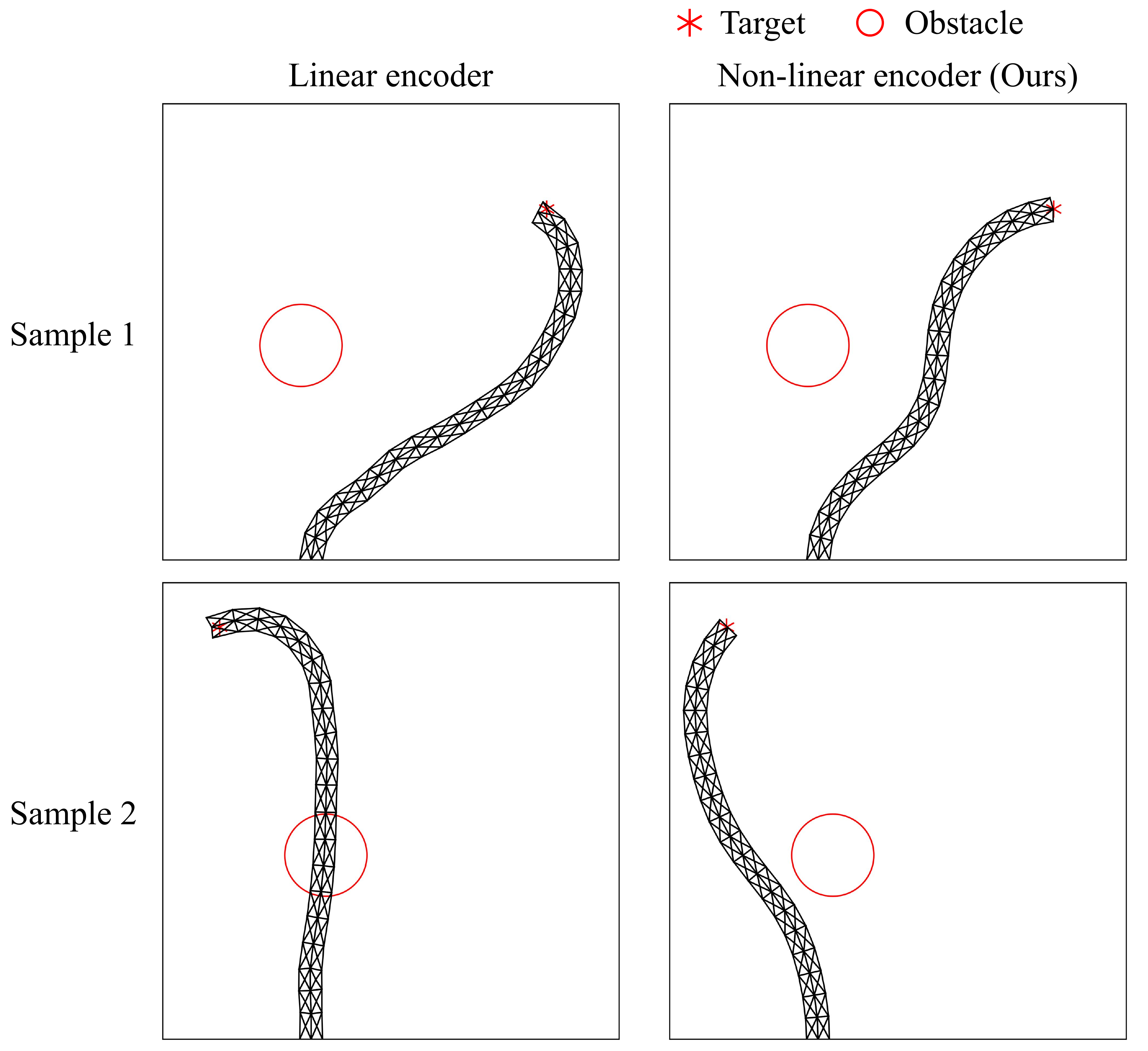}
    \caption{Soft-robot evaluation of linear encoder \textit{vs.} non-linear encoder (ours) on examples from the test set. Linear encoder both violates the constraints (Sample 2) and misses the target (Samples 1 and 2).}
    \label{fig:rb-ablation}
\end{figure*}

%% file: amorsyn.bbl
\begin{thebibliography}{101}
\providecommand{\natexlab}[1]{#1}
\providecommand{\url}[1]{\texttt{#1}}
\expandafter\ifx\csname urlstyle\endcsname\relax
  \providecommand{\doi}[1]{doi: #1}\else
  \providecommand{\doi}{doi: \begingroup \urlstyle{rm}\Url}\fi

\bibitem[Asif(2018)]{asif2018modelling}
Suleman Asif.
\newblock \emph{Modelling and path planning for additive manufacturing of
  continuous fiber composites}.
\newblock PhD thesis, 2018.

\bibitem[Bao et~al.(2020)Bao, Scott, and Sugiyama]{bao2020calibrated}
Han Bao, Clay Scott, and Masashi Sugiyama.
\newblock Calibrated surrogate losses for adversarially robust classification.
\newblock In \emph{Conference on Learning Theory}, pages 408--451. PMLR, 2020.

\bibitem[Barrow et~al.(1977)Barrow, Tenenbaum, Bolles, and
  Wolf]{barrow1977parametric}
Harry~G Barrow, Jay~M Tenenbaum, Robert~C Bolles, and Helen~C Wolf.
\newblock Parametric correspondence and chamfer matching: Two new techniques
  for image matching.
\newblock In \emph{IJCAI}, 1977.

\bibitem[Baykal and Alterovitz(2017)]{baykal2017asymptotically}
Cenk Baykal and Ron Alterovitz.
\newblock Asymptotically optimal design of piecewise cylindrical robots using
  motion planning.
\newblock In \emph{Robotics: Science and Systems}, volume 2017, 2017.

\bibitem[Bendsoe and Sigmund(2013)]{bendsoe2013topology}
Martin~Philip Bendsoe and Ole Sigmund.
\newblock \emph{Topology optimization: theory, methods, and applications}.
\newblock Springer Science \& Business Media, 2013.

\bibitem[Bhosekar and Ierapetritou(2018)]{bhosekar2018advances}
Atharv Bhosekar and Marianthi Ierapetritou.
\newblock Advances in surrogate based modeling, feasibility analysis, and
  optimization: A review.
\newblock \emph{Computers \& Chemical Engineering}, 108:\penalty0 250--267,
  2018.

\bibitem[Bhuvaneswari et~al.(2021)Bhuvaneswari, Priyadharshini, Deepa, Balaji,
  Rajeshkumar, and Ramesh]{bhuvaneswari2021deep}
V~Bhuvaneswari, M~Priyadharshini, C~Deepa, D~Balaji, L~Rajeshkumar, and
  M~Ramesh.
\newblock Deep learning for material synthesis and manufacturing systems: a
  review.
\newblock \emph{Materials Today: Proceedings}, 2021.

\bibitem[Biegler et~al.(2003)Biegler, Ghattas, Heinkenschloss, and van
  Bloemen~Waanders]{biegler2003large}
Lorenz~T Biegler, Omar Ghattas, Matthias Heinkenschloss, and Bart van
  Bloemen~Waanders.
\newblock Large-scale pde-constrained optimization: an introduction.
\newblock In \emph{Large-Scale PDE-Constrained Optimization}, pages 3--13.
  Springer, 2003.

\bibitem[Brookes et~al.(2019)Brookes, Park, and
  Listgarten]{brookes2019conditioning}
David Brookes, Hahnbeom Park, and Jennifer Listgarten.
\newblock Conditioning by adaptive sampling for robust design.
\newblock In \emph{International conference on machine learning}, pages
  773--782. PMLR, 2019.

\bibitem[Brookes and Listgarten(2018)]{brookes2018design}
David~H Brookes and Jennifer Listgarten.
\newblock Design by adaptive sampling.
\newblock \emph{arXiv preprint arXiv:1810.03714}, 2018.

\bibitem[Cabrera et~al.(2002)Cabrera, Simon, and Prado]{cabrera2002optimal}
JA~Cabrera, A~Simon, and M~Prado.
\newblock Optimal synthesis of mechanisms with genetic algorithms.
\newblock \emph{Mechanism and machine theory}, 37\penalty0 (10):\penalty0
  1165--1177, 2002.

\bibitem[Cabrera et~al.(2011)Cabrera, Ortiz, Nadal, and
  Castillo]{cabrera2011evolutionary}
JA~Cabrera, A~Ortiz, F~Nadal, and JJ~Castillo.
\newblock An evolutionary algorithm for path synthesis of mechanisms.
\newblock \emph{Mechanism and Machine Theory}, 46\penalty0 (2):\penalty0
  127--141, 2011.

\bibitem[Campos and Kress-Gazit(2021)]{campos2021synthesizing}
Thais Campos and Hadas Kress-Gazit.
\newblock Synthesizing modular manipulators for tasks with time, obstacle, and
  torque constraints.
\newblock \emph{arXiv preprint arXiv:2106.09487}, 2021.

\bibitem[Campos et~al.(2019)Campos, Inala, Solar-Lezama, and
  Kress-Gazit]{campos2019task}
Thais Campos, Jeevana~Priya Inala, Armando Solar-Lezama, and Hadas Kress-Gazit.
\newblock Task-based design of ad-hoc modular manipulators.
\newblock In \emph{2019 International Conference on Robotics and Automation
  (ICRA)}, pages 6058--6064. IEEE, 2019.

\bibitem[Campos~de Almeida et~al.(2020)Campos~de Almeida, Marri, and
  Kress-Gazit]{campos2020automated}
Thais Campos~de Almeida, Samhita Marri, and Hadas Kress-Gazit.
\newblock Automated synthesis of modular manipulators’ structure and control
  for continuous tasks around obstacles.
\newblock \emph{Robotics: Science and Systems 2020}, 2020.

\bibitem[Chen et~al.(2018)Chen, Engkvist, Wang, Olivecrona, and
  Blaschke]{chen2018rise}
Hongming Chen, Ola Engkvist, Yinhai Wang, Marcus Olivecrona, and Thomas
  Blaschke.
\newblock The rise of deep learning in drug discovery.
\newblock \emph{Drug discovery today}, 23\penalty0 (6):\penalty0 1241--1250,
  2018.

\bibitem[Chen and Burdick(1995)]{chen1995determining}
I-Ming Chen and Joel~W Burdick.
\newblock Determining task optimal modular robot assembly configurations.
\newblock In \emph{proceedings of 1995 IEEE International Conference on
  Robotics and Automation}, volume~1, pages 132--137. IEEE, 1995.

\bibitem[Chung et~al.(1997)Chung, Han, Youm, and Kim]{chung1997task}
Wan~Kyun Chung, Jeongheon Han, Youngil Youm, and SH~Kim.
\newblock Task based design of modular robot manipulator using efficient
  genetic algorithm.
\newblock In \emph{Proceedings of International Conference on Robotics and
  Automation}, volume~1, pages 507--512. IEEE, 1997.

\bibitem[Cianchetti et~al.(2014)Cianchetti, Ranzani, Gerboni, Nanayakkara,
  Althoefer, Dasgupta, and Menciassi]{cianchetti2014soft}
Matteo Cianchetti, Tommaso Ranzani, Giada Gerboni, Thrishantha Nanayakkara,
  Kaspar Althoefer, Prokar Dasgupta, and Arianna Menciassi.
\newblock Soft robotics technologies to address shortcomings in today's
  minimally invasive surgery: the stiff-flop approach.
\newblock \emph{Soft robotics}, 1\penalty0 (2):\penalty0 122--131, 2014.

\bibitem[Coley et~al.(2018)Coley, Green, and Jensen]{coley2018machine}
Connor~W Coley, William~H Green, and Klavs~F Jensen.
\newblock Machine learning in computer-aided synthesis planning.
\newblock \emph{Accounts of chemical research}, 51\penalty0 (5):\penalty0
  1281--1289, 2018.

\bibitem[Coros et~al.(2013)Coros, Thomaszewski, Noris, Sueda, Forberg, Sumner,
  Matusik, and Bickel]{coros2013computational}
Stelian Coros, Bernhard Thomaszewski, Gioacchino Noris, Shinjiro Sueda, Moira
  Forberg, Robert~W Sumner, Wojciech Matusik, and Bernd Bickel.
\newblock Computational design of mechanical characters.
\newblock \emph{ACM Transactions on Graphics (TOG)}, 32\penalty0 (4):\penalty0
  1--12, 2013.

\bibitem[Coumans(2010)]{Coumans2010Bullet}
Erwin Coumans.
\newblock Bullet physics engine.
\newblock \emph{Open Source Software: http://bulletphysics. org}, 2010.

\bibitem[Dogra et~al.(2021)Dogra, Sekhar~Padhee, and Singla]{dogra2021optimal}
Anubhav Dogra, Srikant Sekhar~Padhee, and Ekta Singla.
\newblock An optimal architectural design for unconventional modular
  reconfigurable manipulation system.
\newblock \emph{Journal of Mechanical Design}, 143\penalty0 (6):\penalty0
  063303, 2021.

\bibitem[Engel et~al.(2018)Engel, Hoffman, and Roberts]{engel2018latent}
Jesse Engel, Matthew Hoffman, and Adam Roberts.
\newblock Latent constraints: Learning to generate conditionally from
  unconditional generative models.
\newblock In \emph{International Conference on Learning Representations}, 2018.

\bibitem[Everett et~al.(2018)Everett, Chen, and How]{everett2018motion}
Michael Everett, Yu~Fan Chen, and Jonathan~P How.
\newblock Motion planning among dynamic, decision-making agents with deep
  reinforcement learning.
\newblock In \emph{2018 IEEE/RSJ International Conference on Intelligent Robots
  and Systems (IROS)}, pages 3052--3059. IEEE, 2018.

\bibitem[Fan et~al.(2017)Fan, Su, and Guibas]{fan2017point}
Haoqiang Fan, Hao Su, and Leonidas Guibas.
\newblock A point set generation network for 3d object reconstruction from a
  single image.
\newblock In \emph{CVPR}, 2017.

\bibitem[Fannjiang and Listgarten(2020)]{fannjiang2020autofocused}
Clara Fannjiang and Jennifer Listgarten.
\newblock Autofocused oracles for model-based design.
\newblock \emph{Advances in Neural Information Processing Systems}, 33, 2020.

\bibitem[Farritor et~al.(1996)Farritor, Dubowsky, Rutman, and
  Cole]{farritor1996systems}
Shane Farritor, Steven Dubowsky, Nathan Rutman, and Jeffrey Cole.
\newblock A systems-level modular design approach to field robotics.
\newblock In \emph{Proceedings of IEEE International Conference on Robotics and
  Automation}, volume~4, pages 2890--2895. IEEE, 1996.

\bibitem[Forrester and Keane(2009)]{forrester2009recent}
Alexander~IJ Forrester and Andy~J Keane.
\newblock Recent advances in surrogate-based optimization.
\newblock \emph{Progress in aerospace sciences}, 45\penalty0 (1-3):\penalty0
  50--79, 2009.

\bibitem[Fu and Levine(2020)]{fu2020offline}
Justin Fu and Sergey Levine.
\newblock Offline model-based optimization via normalized maximum likelihood
  estimation.
\newblock In \emph{International Conference on Learning Representations}, 2020.

\bibitem[Ganganath et~al.(2016)Ganganath, Cheng, Fok, and
  Chi]{ganganath2016trajectory}
Nuwan Ganganath, Chi-Tsun Cheng, Kai-Yin Fok, and K~Tse Chi.
\newblock Trajectory planning for 3d printing: A revisit to traveling salesman
  problem.
\newblock In \emph{2016 2nd International Conference on Control, Automation and
  Robotics (ICCAR)}, pages 287--290. IEEE, 2016.

\bibitem[G{\'o}mez-Bombarelli et~al.(2018)G{\'o}mez-Bombarelli, Wei, Duvenaud,
  Hern{\'a}ndez-Lobato, S{\'a}nchez-Lengeling, Sheberla, Aguilera-Iparraguirre,
  Hirzel, Adams, and Aspuru-Guzik]{gomez2018automatic}
Rafael G{\'o}mez-Bombarelli, Jennifer~N Wei, David Duvenaud, Jos{\'e}~Miguel
  Hern{\'a}ndez-Lobato, Benjam{\'\i}n S{\'a}nchez-Lengeling, Dennis Sheberla,
  Jorge Aguilera-Iparraguirre, Timothy~D Hirzel, Ryan~P Adams, and Al{\'a}n
  Aspuru-Guzik.
\newblock Automatic chemical design using a data-driven continuous
  representation of molecules.
\newblock \emph{ACS central science}, 4\penalty0 (2):\penalty0 268--276, 2018.

\bibitem[Grabocka et~al.(2019)Grabocka, Scholz, and
  Schmidt-Thieme]{grabocka2019learning}
Josif Grabocka, Randolf Scholz, and Lars Schmidt-Thieme.
\newblock Learning surrogate losses.
\newblock \emph{arXiv preprint arXiv:1905.10108}, 2019.

\bibitem[Guan et~al.(2020)Guan, Liu, Liu, Yin, Hu, van Kaick, Zhang, Yumer,
  Carr, Mech, and Zhang]{guan2020fame}
Yanran Guan, Han Liu, Kun Liu, Kangxue Yin, Ruizhen Hu, Oliver van Kaick, Yan
  Zhang, Ersin Yumer, Nathan Carr, Radomir Mech, and Hao Zhang.
\newblock {FAME}: 3d shape generation via functionality-aware model evolution.
\newblock \emph{IEEE Transactions on Visualization and Computer Graphics},
  2020.

\bibitem[Gulwani(2014)]{gulwani2014program}
Sumit Gulwani.
\newblock Program synthesis.
\newblock \emph{Software Systems Safety}, pages 43--75, 2014.

\bibitem[Gupta and Zou(2019)]{gupta2019feedback}
Anvita Gupta and James Zou.
\newblock Feedback gan for dna optimizes protein functions.
\newblock \emph{Nature Machine Intelligence}, 1\penalty0 (2):\penalty0
  105--111, 2019.

\bibitem[Ha et~al.(2016)Ha, Coros, Alspach, Kim, and Yamane]{ha2016task}
Sehoon Ha, Stelian Coros, Alexander Alspach, Joohyung Kim, and Katsu Yamane.
\newblock Task-based limb optimization for legged robots.
\newblock In \emph{2016 IEEE/RSJ International Conference on Intelligent Robots
  and Systems (IROS)}, pages 2062--2068. IEEE, 2016.

\bibitem[Ha et~al.(2018)Ha, Coros, Alspach, Bern, Kim, and
  Yamane]{ha2018computational}
Sehoon Ha, Stelian Coros, Alexander Alspach, James~M Bern, Joohyung Kim, and
  Katsu Yamane.
\newblock Computational design of robotic devices from high-level motion
  specifications.
\newblock \emph{IEEE Transactions on Robotics}, 34\penalty0 (5):\penalty0
  1240--1251, 2018.

\bibitem[Han et~al.(2012)Han, Zhang, et~al.]{han2012surrogate}
Zhong-Hua Han, Ke-Shi Zhang, et~al.
\newblock Surrogate-based optimization.
\newblock \emph{Real-world applications of genetic algorithms}, 343, 2012.

\bibitem[Hanneke et~al.(2019)Hanneke, Yang, et~al.]{hanneke2019surrogate}
Steve Hanneke, Liu Yang, et~al.
\newblock Surrogate losses in passive and active learning.
\newblock \emph{Electronic Journal of Statistics}, 13\penalty0 (2):\penalty0
  4646--4708, 2019.

\bibitem[Hawkins-Hooker et~al.(2021)Hawkins-Hooker, Depardieu, Baur, Couairon,
  Chen, and Bikard]{hawkins2021generating}
Alex Hawkins-Hooker, Florence Depardieu, Sebastien Baur, Guillaume Couairon,
  Arthur Chen, and David Bikard.
\newblock Generating functional protein variants with variational autoencoders.
\newblock \emph{PLoS computational biology}, 17\penalty0 (2):\penalty0
  e1008736, 2021.

\bibitem[Hornby et~al.(2001)Hornby, Lipson, and Pollack]{hornby2001evolution}
Gregory~S Hornby, Hod Lipson, and Jordan~B Pollack.
\newblock Evolution of generative design systems for modular physical robots.
\newblock In \emph{Proceedings 2001 ICRA. IEEE International Conference on
  Robotics and Automation (Cat. No. 01CH37164)}, volume~4, pages 4146--4151.
  IEEE, 2001.

\bibitem[Huang et~al.(2020)Huang, Chen, Jiang, Zou, Li, Liu, and
  Yu]{huang2020survey}
Jiaqi Huang, Qian Chen, Hao Jiang, Bin Zou, Lei Li, Jikai Liu, and Huangchao
  Yu.
\newblock A survey of design methods for material extrusion polymer 3d
  printing.
\newblock \emph{Virtual and Physical Prototyping}, 15\penalty0 (2):\penalty0
  148--162, 2020.

\bibitem[Hughes(2012)]{hughes2012finite}
Thomas~JR Hughes.
\newblock \emph{The finite element method: linear static and dynamic finite
  element analysis}.
\newblock Courier Corporation, 2012.

\bibitem[Huo et~al.(2019)Huo, Rong, Kononova, Sun, Botari, He, Tshitoyan, and
  Ceder]{huo2019semi}
Haoyan Huo, Ziqin Rong, Olga Kononova, Wenhao Sun, Tiago Botari, Tanjin He,
  Vahe Tshitoyan, and Gerbrand Ceder.
\newblock Semi-supervised machine-learning classification of materials
  synthesis procedures.
\newblock \emph{npj Computational Materials}, 5\penalty0 (1):\penalty0 1--7,
  2019.

\bibitem[Jin et~al.(2020)Jin, Barzilay, and Jaakkola]{jin2020multi}
Wengong Jin, Regina Barzilay, and Tommi Jaakkola.
\newblock Multi-objective molecule generation using interpretable
  substructures.
\newblock In \emph{International Conference on Machine Learning}, pages
  4849--4859. PMLR, 2020.

\bibitem[Jin et~al.(2021)Jin, Stokes, Eastman, Itkin, Zakharov, Collins,
  Jaakkola, and Barzilay]{jin2021deep}
Wengong Jin, Jonathan~M Stokes, Richard~T Eastman, Zina Itkin, Alexey~V
  Zakharov, James~J Collins, Tommi~S Jaakkola, and Regina Barzilay.
\newblock Deep learning identifies synergistic drug combinations for treating
  covid-19.
\newblock \emph{Proceedings of the National Academy of Sciences}, 118\penalty0
  (39), 2021.

\bibitem[Kallioras et~al.(2020)Kallioras, Kazakis, and
  Lagaros]{kallioras2020accelerated}
Nikos~Ath Kallioras, Georgios Kazakis, and Nikos~D Lagaros.
\newblock Accelerated topology optimization by means of deep learning.
\newblock \emph{Structural and Multidisciplinary Optimization}, 62\penalty0
  (3):\penalty0 1185--1212, 2020.

\bibitem[Karaman and Frazzoli(2011)]{karaman2011sampling}
Sertac Karaman and Emilio Frazzoli.
\newblock Sampling-based algorithms for optimal motion planning.
\newblock \emph{The international journal of robotics research}, 30\penalty0
  (7):\penalty0 846--894, 2011.

\bibitem[Khorshidi et~al.(2011)Khorshidi, Soheilypour, Peyro, Atai, and
  Panahi]{khorshidi2011optimal}
M~Khorshidi, M~Soheilypour, M~Peyro, A~Atai, and M~Shariat Panahi.
\newblock Optimal design of four-bar mechanisms using a hybrid multi-objective
  ga with adaptive local search.
\newblock \emph{Mechanism and Machine Theory}, 46\penalty0 (10):\penalty0
  1453--1465, 2011.

\bibitem[Killoran et~al.(2017)Killoran, Lee, Delong, Duvenaud, and
  Frey]{killoran2017generating}
Nathan Killoran, Leo~J Lee, Andrew Delong, David Duvenaud, and Brendan~J Frey.
\newblock Generating and designing dna with deep generative models.
\newblock \emph{arXiv preprint arXiv:1712.06148}, 2017.

\bibitem[Kim et~al.(2017)Kim, Huang, Saunders, McCallum, Ceder, and
  Olivetti]{kim2017materials}
Edward Kim, Kevin Huang, Adam Saunders, Andrew McCallum, Gerbrand Ceder, and
  Elsa Olivetti.
\newblock Materials synthesis insights from scientific literature via text
  extraction and machine learning.
\newblock \emph{Chemistry of Materials}, 29\penalty0 (21):\penalty0 9436--9444,
  2017.

\bibitem[Kim and Khosla(1993)]{kim1993formulation}
J-O Kim and Pradeep~K Khosla.
\newblock A formulation for task based design of robot manipulators.
\newblock In \emph{Proceedings of 1993 IEEE/RSJ International Conference on
  Intelligent Robots and Systems (IROS'93)}, volume~3, pages 2310--2317. IEEE,
  1993.

\bibitem[Kingma and Ba(2015)]{Kingma2015Adam:}
Diederik~P. Kingma and Jimmy Ba.
\newblock Adam: A method for stochastic optimization.
\newblock In \emph{ICLR}, 2015.

\bibitem[Kollmann et~al.(2020)Kollmann, Abueidda, Koric, Guleryuz, and
  Sobh]{kollmann2020deep}
Hunter~T Kollmann, Diab~W Abueidda, Seid Koric, Erman Guleryuz, and Nahil~A
  Sobh.
\newblock Deep learning for topology optimization of 2d metamaterials.
\newblock \emph{Materials \& Design}, 196:\penalty0 109098, 2020.

\bibitem[Koziel and Leifsson(2013)]{koziel2013surrogate}
Slawomir Koziel and Leifur Leifsson.
\newblock \emph{Surrogate-based modeling and optimization}.
\newblock Springer, 2013.

\bibitem[Koziel and Ogurtsov(2014)]{koziel2014surrogate}
Slawomir Koziel and Stanislav Ogurtsov.
\newblock Surrogate-based optimization.
\newblock In \emph{Antenna Design by Simulation-Driven Optimization}, pages
  13--24. Springer, 2014.

\bibitem[Koziel et~al.(2011)Koziel, Ciaurri, and Leifsson]{koziel2011surrogate}
Slawomir Koziel, David~Echeverr{\'\i}a Ciaurri, and Leifur Leifsson.
\newblock Surrogate-based methods.
\newblock In \emph{Computational optimization, methods and algorithms}, pages
  33--59. Springer, 2011.

\bibitem[Latombe(2012)]{latombe2012robot}
Jean-Claude Latombe.
\newblock \emph{Robot motion planning}, volume 124.
\newblock Springer Science \& Business Media, 2012.

\bibitem[Leger and Bares(1999)]{leger1999automated}
Chris Leger and John Bares.
\newblock Automated task-based synthesis and optimization of field robots.
\newblock 1999.

\bibitem[Li et~al.(2021)Li, Chen, Zhou, Liang, Xiao, Cao, Zhang, Zhang, Wu,
  Yin, et~al.]{li2021self}
Guorui Li, Xiangping Chen, Fanghao Zhou, Yiming Liang, Youhua Xiao, Xunuo Cao,
  Zhen Zhang, Mingqi Zhang, Baosheng Wu, Shunyu Yin, et~al.
\newblock Self-powered soft robot in the mariana trench.
\newblock \emph{Nature}, 591\penalty0 (7848):\penalty0 66--71, 2021.

\bibitem[Liu et~al.(2020)Liu, Wang, and Deng]{liu2020unified}
Lanlan Liu, Mingzhe Wang, and Jia Deng.
\newblock A unified framework of surrogate loss by refactoring and
  interpolation.
\newblock In \emph{ECCV}, 2020.

\bibitem[Murray et~al.(2010)Murray, Adams, and MacKay]{murray2010elliptical}
Iain Murray, Ryan Adams, and David MacKay.
\newblock Elliptical slice sampling.
\newblock In \emph{Proceedings of the thirteenth international conference on
  artificial intelligence and statistics}, pages 541--548. JMLR Workshop and
  Conference Proceedings, 2010.

\bibitem[Nagendar et~al.(2018)Nagendar, Singh, Balasubramanian, and
  Jawahar]{nagendar2018neuro}
Gattigorla Nagendar, Digvijay Singh, Vineeth~N Balasubramanian, and CV~Jawahar.
\newblock Neuro-iou: Learning a surrogate loss for semantic segmentation.
\newblock In \emph{BMVC}, page 278, 2018.

\bibitem[Nesterov et~al.(2018)]{nesterov2018lectures}
Yurii Nesterov et~al.
\newblock \emph{Lectures on convex optimization}, volume 137.
\newblock Springer, 2018.

\bibitem[Nocedal and Wright(2006)]{nocedal2006numerical}
Jorge Nocedal and Stephen Wright.
\newblock \emph{Numerical optimization}.
\newblock Springer Science \& Business Media, 2006.

\bibitem[Pang et~al.(2020)Pang, Yang, Heng, Ye, Huang, Yang, and
  Pang]{pang2020coboskin}
Gaoyang Pang, Geng Yang, Wenzheng Heng, Zhiqiu Ye, Xiaoyan Huang, Hua-Yong
  Yang, and Zhibo Pang.
\newblock Coboskin: Soft robot skin with variable stiffness for safer
  human--robot collaboration.
\newblock \emph{IEEE Transactions on Industrial Electronics}, 68\penalty0
  (4):\penalty0 3303--3314, 2020.

\bibitem[Paszke et~al.(2019)Paszke, Gross, Massa, Lerer, Bradbury, Chanan,
  Killeen, Lin, Gimelshein, Antiga, et~al.]{paszke2019pytorch}
Adam Paszke, Sam Gross, Francisco Massa, Adam Lerer, James Bradbury, Gregory
  Chanan, Trevor Killeen, Zeming Lin, Natalia Gimelshein, Luca Antiga, et~al.
\newblock Pytorch: An imperative style, high-performance deep learning library.
\newblock In \emph{Advances in neural information processing systems}, pages
  8026--8037, 2019.

\bibitem[Patel and Sobh(2015)]{patel2015task}
Sarosh Patel and Tarek Sobh.
\newblock Task based synthesis of serial manipulators.
\newblock \emph{Journal of advanced research}, 6\penalty0 (3):\penalty0
  479--492, 2015.

\bibitem[Patel et~al.(2020)Patel, Hoda{\v{n}}, and Matas]{patel2020learning}
Yash Patel, Tom{\'a}{\v{s}} Hoda{\v{n}}, and Ji{\v{r}}{\'\i} Matas.
\newblock Learning surrogates via deep embedding.
\newblock In \emph{European Conference on Computer Vision}, pages 205--221.
  Springer, 2020.

\bibitem[Polykovskiy et~al.(2018)Polykovskiy, Zhebrak, Vetrov, Ivanenkov,
  Aladinskiy, Mamoshina, Bozdaganyan, Aliper, Zhavoronkov, and
  Kadurin]{polykovskiy2018entangled}
Daniil Polykovskiy, Alexander Zhebrak, Dmitry Vetrov, Yan Ivanenkov, Vladimir
  Aladinskiy, Polina Mamoshina, Marine Bozdaganyan, Alexander Aliper, Alex
  Zhavoronkov, and Artur Kadurin.
\newblock Entangled conditional adversarial autoencoder for de novo drug
  discovery.
\newblock \emph{Molecular pharmaceutics}, 15\penalty0 (10):\penalty0
  4398--4405, 2018.

\bibitem[Renda et~al.(2020)Renda, Chen, Mendis, and Carbin]{renda2020difftune}
Alex Renda, Yishen Chen, Charith Mendis, and Michael Carbin.
\newblock Difftune: Optimizing cpu simulator parameters with learned
  differentiable surrogates.
\newblock In \emph{2020 53rd Annual IEEE/ACM International Symposium on
  Microarchitecture (MICRO)}, pages 442--455. IEEE, 2020.

\bibitem[Rives et~al.(2021)Rives, Meier, Sercu, Goyal, Lin, Liu, Guo, Ott,
  Zitnick, Ma, et~al.]{rives2021biological}
Alexander Rives, Joshua Meier, Tom Sercu, Siddharth Goyal, Zeming Lin, Jason
  Liu, Demi Guo, Myle Ott, C~Lawrence Zitnick, Jerry Ma, et~al.
\newblock Biological structure and function emerge from scaling unsupervised
  learning to 250 million protein sequences.
\newblock \emph{Proceedings of the National Academy of Sciences}, 118\penalty0
  (15), 2021.

\bibitem[Rus and Tolley(2015)]{rus2015design}
Daniela Rus and Michael~T Tolley.
\newblock Design, fabrication and control of soft robots.
\newblock \emph{Nature}, 521\penalty0 (7553):\penalty0 467--475, 2015.

\bibitem[Sasaki and Igarashi(2019)]{sasaki2019topology}
Hidenori Sasaki and Hajime Igarashi.
\newblock Topology optimization accelerated by deep learning.
\newblock \emph{IEEE Transactions on Magnetics}, 55\penalty0 (6):\penalty0
  1--5, 2019.

\bibitem[Seff et~al.(2019)Seff, Zhou, Damani, Doyle, and
  Adams]{NEURIPS2019_febefe1c}
Ari Seff, Wenda Zhou, Farhan Damani, Abigail Doyle, and Ryan~P Adams.
\newblock Discrete object generation with reversible inductive construction.
\newblock In \emph{Advances in Neural Information Processing Systems},
  volume~32, pages 10353--10363, 2019.

\bibitem[Shahrubudin et~al.(2019)Shahrubudin, Lee, and
  Ramlan]{shahrubudin2019overview}
Nurhalida Shahrubudin, Te~Chuan Lee, and Rhaizan Ramlan.
\newblock An overview on 3d printing technology: technological, materials, and
  applications.
\newblock \emph{Procedia Manufacturing}, 35:\penalty0 1286--1296, 2019.

\bibitem[Shembekar et~al.(2018)Shembekar, Yoon, Kanyuck, and
  Gupta]{shembekar2018trajectory}
Aniruddha~V Shembekar, Yeo~Jung Yoon, Alec Kanyuck, and Satyandra~K Gupta.
\newblock Trajectory planning for conformal 3d printing using non-planar
  layers.
\newblock In \emph{International Design Engineering Technical Conferences and
  Computers and Information in Engineering Conference}, volume 51722, page
  V01AT02A026. American Society of Mechanical Engineers, 2018.

\bibitem[Shirobokov et~al.(2020)Shirobokov, Belavin, Kagan, Ustyuzhanin, and
  Baydin]{shirobokov2020black}
Sergey Shirobokov, Vladislav Belavin, Michael Kagan, Andrei Ustyuzhanin, and
  Atilim~Gunes Baydin.
\newblock Black-box optimization with local generative surrogates.
\newblock In \emph{Workshop on Real World Experiment Design and Active Learning
  at International Conference on Machine Learning}, 2020.

\bibitem[Sigmund and Maute(2013)]{sigmund2013topology}
Ole Sigmund and Kurt Maute.
\newblock Topology optimization approaches.
\newblock \emph{Structural and Multidisciplinary Optimization}, 48\penalty0
  (6):\penalty0 1031--1055, 2013.

\bibitem[Singh et~al.(1994)Singh, Barto, Grupen, Connolly,
  et~al.]{singh1994robust}
Satinder~P Singh, Andrew~G Barto, Roderic Grupen, Christopher Connolly, et~al.
\newblock Robust reinforcement learning in motion planning.
\newblock \emph{Advances in neural information processing systems}, pages
  655--655, 1994.

\bibitem[Sleesongsom and Bureerat(2017)]{sleesongsom2017four}
Suwin Sleesongsom and Sujin Bureerat.
\newblock Four-bar linkage path generation through self-adaptive population
  size teaching-learning based optimization.
\newblock \emph{Knowledge-Based Systems}, 135:\penalty0 180--191, 2017.

\bibitem[Stokes et~al.(2020)Stokes, Yang, Swanson, Jin, Cubillos-Ruiz, Donghia,
  MacNair, French, Carfrae, Bloom-Ackermann, et~al.]{stokes2020deep}
Jonathan~M Stokes, Kevin Yang, Kyle Swanson, Wengong Jin, Andres Cubillos-Ruiz,
  Nina~M Donghia, Craig~R MacNair, Shawn French, Lindsey~A Carfrae, Zohar
  Bloom-Ackermann, et~al.
\newblock A deep learning approach to antibiotic discovery.
\newblock \emph{Cell}, 180\penalty0 (4):\penalty0 688--702, 2020.

\bibitem[Stragiotti(2020)]{stragiotti2020continuous}
Enrico Stragiotti.
\newblock \emph{Continuous Fiber Path Planning Algorithm for 3D Printed Optimal
  Mechanical Properties.}
\newblock PhD thesis, Politecnico di Torino, 2020.

\bibitem[Subramanian et~al.(1995)]{subramanian1995kinematic}
Devika Subramanian et~al.
\newblock Kinematic synthesis with configuration spaces.
\newblock \emph{Research in Engineering Design}, 7\penalty0 (3):\penalty0
  193--213, 1995.

\bibitem[Sun et~al.(2018)Sun, Wu, Zhang, Zhang, Zhang, Xue, Tenenbaum, and
  Freeman]{pix3d}
Xingyuan Sun, Jiajun Wu, Xiuming Zhang, Zhoutong Zhang, Chengkai Zhang, Tianfan
  Xue, Joshua~B Tenenbaum, and William~T Freeman.
\newblock Pix3d: Dataset and methods for single-image 3d shape modeling.
\newblock In \emph{IEEE Conference on Computer Vision and Pattern Recognition
  (CVPR)}, 2018.

\bibitem[Tabandeh et~al.(2016)Tabandeh, Melek, Biglarbegian, Won, and
  Clark]{tabandeh2016memetic}
Saleh Tabandeh, William Melek, Mohammad Biglarbegian, Seong-hoon~Peter Won, and
  Chris Clark.
\newblock A memetic algorithm approach for solving the task-based configuration
  optimization problem in serial modular and reconfigurable robots.
\newblock \emph{Robotica}, 34\penalty0 (9):\penalty0 1979--2008, 2016.

\bibitem[Tian et~al.(2018)Tian, Luo, Sun, Ellis, Freeman, Tenenbaum, and
  Wu]{tian2018learning}
Yonglong Tian, Andrew Luo, Xingyuan Sun, Kevin Ellis, William~T Freeman,
  Joshua~B Tenenbaum, and Jiajun Wu.
\newblock Learning to infer and execute 3d shape programs.
\newblock In \emph{International Conference on Learning Representations}, 2018.

\bibitem[Trabucco et~al.(2021)Trabucco, Kumar, Geng, and
  Levine]{trabucco2021conservative}
Brandon Trabucco, Aviral Kumar, Xinyang Geng, and Sergey Levine.
\newblock Conservative objective models for effective offline model-based
  optimization.
\newblock In \emph{International Conference on Machine Learning}, pages
  10358--10368. PMLR, 2021.

\bibitem[Vamathevan et~al.(2019)Vamathevan, Clark, Czodrowski, Dunham, Ferran,
  Lee, Li, Madabhushi, Shah, Spitzer, et~al.]{vamathevan2019applications}
Jessica Vamathevan, Dominic Clark, Paul Czodrowski, Ian Dunham, Edgardo Ferran,
  George Lee, Bin Li, Anant Madabhushi, Parantu Shah, Michaela Spitzer, et~al.
\newblock Applications of machine learning in drug discovery and development.
\newblock \emph{Nature Reviews Drug Discovery}, 18\penalty0 (6):\penalty0
  463--477, 2019.

\bibitem[Van~Henten et~al.(2009)Van~Henten, Van’t~Slot, Hol, and
  Van~Willigenburg]{van2009optimal}
EJ~Van~Henten, DA~Van’t~Slot, CWJ Hol, and LG~Van~Willigenburg.
\newblock Optimal manipulator design for a cucumber harvesting robot.
\newblock \emph{Computers and electronics in agriculture}, 65\penalty0
  (2):\penalty0 247--257, 2009.

\bibitem[Virtanen et~al.(2020)Virtanen, Gommers, Oliphant, Haberland, Reddy,
  Cournapeau, Burovski, Peterson, Weckesser, Bright, {van der Walt}, Brett,
  Wilson, Millman, Mayorov, Nelson, Jones, Kern, Larson, Carey, Polat, Feng,
  Moore, {VanderPlas}, Laxalde, Perktold, Cimrman, Henriksen, Quintero, Harris,
  Archibald, Ribeiro, Pedregosa, {van Mulbregt}, and {SciPy 1.0
  Contributors}]{2020SciPyNMeth}
Pauli Virtanen, Ralf Gommers, Travis~E. Oliphant, Matt Haberland, Tyler Reddy,
  David Cournapeau, Evgeni Burovski, Pearu Peterson, Warren Weckesser, Jonathan
  Bright, St{\'e}fan~J. {van der Walt}, Matthew Brett, Joshua Wilson, K.~Jarrod
  Millman, Nikolay Mayorov, Andrew R.~J. Nelson, Eric Jones, Robert Kern, Eric
  Larson, C~J Carey, {\.I}lhan Polat, Yu~Feng, Eric~W. Moore, Jake
  {VanderPlas}, Denis Laxalde, Josef Perktold, Robert Cimrman, Ian Henriksen,
  E.~A. Quintero, Charles~R. Harris, Anne~M. Archibald, Ant{\^o}nio~H. Ribeiro,
  Fabian Pedregosa, Paul {van Mulbregt}, and {SciPy 1.0 Contributors}.
\newblock {{SciPy} 1.0: Fundamental Algorithms for Scientific Computing in
  Python}.
\newblock \emph{Nature Methods}, 17:\penalty0 261--272, 2020.
\newblock \doi{10.1038/s41592-019-0686-2}.

\bibitem[Wang et~al.(2003)Wang, Wang, and Guo]{wang2003level}
Michael~Yu Wang, Xiaoming Wang, and Dongming Guo.
\newblock A level set method for structural topology optimization.
\newblock \emph{Computer methods in applied mechanics and engineering},
  192\penalty0 (1-2):\penalty0 227--246, 2003.

\bibitem[Whitman and Choset(2018)]{whitman2018task}
Julian Whitman and Howie Choset.
\newblock Task-specific manipulator design and trajectory synthesis.
\newblock \emph{IEEE Robotics and Automation Letters}, 4\penalty0 (2):\penalty0
  301--308, 2018.

\bibitem[Wu et~al.(2017)Wu, Wang, Xue, Sun, Freeman, and Tenenbaum]{marrnet}
Jiajun Wu, Yifan Wang, Tianfan Xue, Xingyuan Sun, William~T Freeman, and
  Joshua~B Tenenbaum.
\newblock {MarrNet: 3D Shape Reconstruction via 2.5D Sketches}.
\newblock In \emph{NIPS}, 2017.

\bibitem[Xiao et~al.(2020)Xiao, Han, Tang, and Duan]{xiao2020efficient}
Hong Xiao, Wei Han, Wenbin Tang, and Yugang Duan.
\newblock An efficient and adaptable path planning algorithm for automated
  fiber placement based on meshing and multi guidelines.
\newblock \emph{Materials}, 13\penalty0 (18):\penalty0 4209, 2020.

\bibitem[Xue et~al.(2020)Xue, Beatson, Adriaenssens, and
  Adams]{xue2020amortized}
Tianju Xue, Alex Beatson, Sigrid Adriaenssens, and Ryan Adams.
\newblock Amortized finite element analysis for fast pde-constrained
  optimization.
\newblock In \emph{International Conference on Machine Learning}, pages
  10638--10647. PMLR, 2020.

\bibitem[Yi et~al.(2018)Yi, Wu, Gan, Torralba, Kohli, and
  Tenenbaum]{yi2018neural}
Kexin Yi, Jiajun Wu, Chuang Gan, Antonio Torralba, Pushmeet Kohli, and Joshua~B
  Tenenbaum.
\newblock Neural-symbolic vqa: disentangling reasoning from vision and language
  understanding.
\newblock In \emph{Proceedings of the 32nd International Conference on Neural
  Information Processing Systems}, pages 1039--1050, 2018.

\bibitem[Yu et~al.(2018)Yu, Hur, and Jung]{yu2018deep}
Yonggyun Yu, Taeil Hur, and Jaeho Jung.
\newblock Deep learning for topology optimization design.
\newblock \emph{arXiv preprint arXiv:1801.05463}, 2018.

\bibitem[Yuan et~al.(2020)Yuan, Yan, Sonka, and Yang]{yuan2020robust}
Zhuoning Yuan, Yan Yan, Milan Sonka, and Tianbao Yang.
\newblock Robust deep auc maximization: A new surrogate loss and empirical
  studies on medical image classification.
\newblock \emph{arXiv preprint arXiv:2012.03173}, 2020.

\bibitem[Zhu et~al.(2012)Zhu, Xu, Snyder, Liu, Wang, and Guo]{zhu2012motion}
Lifeng Zhu, Weiwei Xu, John Snyder, Yang Liu, Guoping Wang, and Baining Guo.
\newblock Motion-guided mechanical toy modeling.
\newblock \emph{ACM Transactions on Graphics (TOG)}, 31\penalty0 (6):\penalty0
  1--10, 2012.

\end{thebibliography}
